\documentclass[10pt,twocolumn,letterpaper]{article}

\usepackage{cvpr}              

\usepackage{microtype}

\usepackage{amsmath}
\usepackage{booktabs}
\usepackage{graphicx}
\usepackage{multirow}

\definecolor{cvprblue}{rgb}{0.21,0.49,0.74}
\usepackage[pagebackref,breaklinks,colorlinks,allcolors=cvprblue]{hyperref}

\def\paperID{REDACTED} 
\def\confName{CVPR}
\def\confYear{2026}

\title{TRGS-SLAM: IMU-Aided Gaussian Splatting SLAM for Blurry, Rolling Shutter, and Noisy Thermal Images}

\author{Spencer Carmichael\textsuperscript{\dag}, Katherine A. Skinner\\
University of Michigan, Ann Arbor, MI USA\\
\tt\small \{specarmi,kskin\}@umich.edu
}

\begin{document}
\maketitle
\begin{abstract}
Thermal cameras offer several advantages for simultaneous localization and mapping (SLAM) with mobile robots: they provide a passive, low-power solution to operating in darkness, are invariant to rapidly changing or high dynamic range illumination, and can see through fog, dust, and smoke. However, uncooled microbolometer thermal cameras, the only practical option in most robotics applications, suffer from significant motion blur, rolling shutter distortions, and fixed pattern noise. In this paper, we present TRGS-SLAM, a 3D Gaussian Splatting (3DGS) based thermal inertial SLAM system uniquely capable of handling these degradations. To overcome the challenges of thermal data, we introduce a model-aware 3DGS rendering method and several general innovations to 3DGS SLAM, including B-spline trajectory optimization with a two-stage IMU loss, view-diversity-based opacity resetting, and pose drift correction schemes. Our system demonstrates accurate tracking on real-world, fast motion, and high-noise thermal data that causes all other tested SLAM methods to fail. Moreover, through offline refinement of our SLAM results, we demonstrate thermal image restoration competitive with prior work that required ground truth poses. The code is available at: \url{https://umautobots.github.io/trgs_slam}.
\end{abstract}    
{
\let\thefootnote\relax\footnotetext{\textsuperscript{\dag}Corresponding author.}
}
\vspace{-5mm}\section{Introduction} 

\begin{figure}
\centering
\includegraphics[width=1.0\columnwidth]{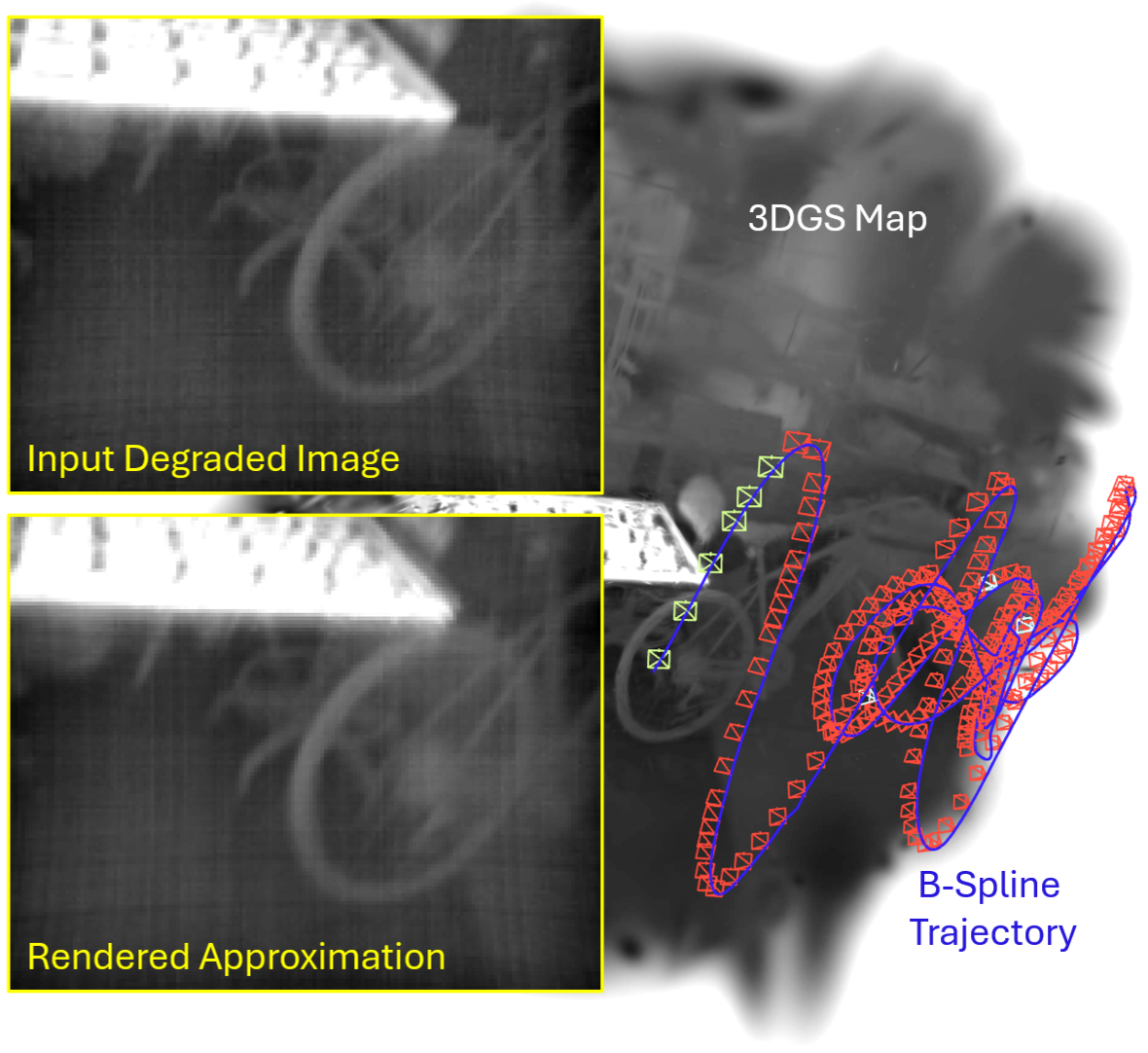}
\caption{The main elements of TRGS-SLAM. \textbf{Left:} Microbolometer-aware rendering. \textbf{Right:} 3DGS map and B-spline trajectory.\vspace{-5mm}}  
\label{figure:intro_image}
\end{figure}

The capability of thermal cameras to operate under adverse illumination and see through visual obscurants opens new applications for simultaneous localization and mapping (SLAM) frameworks for mobile robots. A highly motivating example is a thermal camera-equipped robot mapping a dangerous environment, bringing valuable situational awareness to rescue workers \cite{van_manen_firebotslam_2023}. Given the low weight, power consumption, and price of uncooled microbolometer thermal cameras (relative to their cooled counterparts \cite{yadav_advancements_2022, gade_thermal_2014}), they are the likely choice in this scenario. However, these cameras pose challenges for SLAM. Under slow motion, the low contrast and fixed pattern noise (FPN) of the images are the main issues, and these topics have been well addressed \cite{zhao_tp-tio_2020, lee2024self,wang_edge-based_2023,chen2024eamt,hou2025robust,lai_tpl-slam_2025,polizzi2022data,wu2025monocular}. However, in some scenarios, including the search and rescue context, aggressive camera motion may be necessary. In this case, the images are further afflicted by motion blur and rolling shutter distortions, problems that are seldom addressed in prior work on thermal SLAM.

Recent work, TRNeRF, has demonstrated that given ground truth poses, thermal images with significant FPN, motion blur, and rolling shutter distortions can be restored \cite{specarmi_trnerf}. This is accomplished by modeling microbolometer degradations within the rendering pipeline of a Neural Radiance Field (NeRF) \cite{mildenhall_nerf_2020}. Concurrently, in the visual domain, related techniques on motion deblurring with NeRF and 3D Gaussian Splatting (3DGS) \cite{wang_bad-nerf_2023,lee_exblurf_2023,zhao_bad-gaussians_2024, chen_deblur-gs_2024} have been incorporated in online SLAM systems \cite{bae_i2-slam_2024,wang_mba-slam_2024,girlanda2025deblur}.

In this paper, we extend the ideas of TRNeRF into an online 3DGS SLAM system (\cref{figure:intro_image}). This requires significant innovation, as the frontend tracking systems and monocular depth networks relied on by many NeRF and 3DGS SLAM systems are unreliable under the degradations we seek to overcome. Instead, we utilize a B-spline trajectory and tightly fuse with an IMU to estimate and constrain the continuous poses needed to model blur and rolling shutter. We also introduce other general ideas to 3DGS SLAM, including new techniques for opacity resetting and pose drift correction. We demonstrate, with a thorough benchmark on the TRNeRF dataset, that our method is uniquely capable of accurate tracking under severe thermal degradations. In addition, through offline refinement of our SLAM results, we achieve restoration performance comparable to TRNeRF, which utilizes ground truth poses from stereo visual structure from motion.

In summary, our main contributions are as follows:
\begin{itemize}
    \item A thermal inertial SLAM system, with a 3DGS map and continuous time B-spline trajectory, uniquely capable of accurate tracking under severe image degradation.
    \item Novel 3DGS SLAM methods, including microbolometer-aware rendering, a two-stage direct IMU loss, view-diversity-based opacity resetting, and pose drift correction schemes. 
    \item A system for offline refinement of the SLAM results that enables thermal image restoration competitive with prior work, but without the need for ground truth poses.
\end{itemize}

\section{Related Work}

\paragraph{Thermal SLAM}

Existing work on thermal SLAM has primarily focused on the low contrast and high noise often present in thermal imagery. Significant effort has been made to identify features robust to these issues: handcrafted features have been exhaustively benchmarked \cite{vidas_exploration_2011,mouats_performance_2018,van_manen_firebotslam_2023}, learned features have been trained with FPN-based data augmentation \cite{zhao_tp-tio_2020, lee2024self}, and edge- \cite{wang_edge-based_2023,chen2024eamt,hou2025robust} or line- \cite{lai_tpl-slam_2025} based features have been investigated. Rather than relying on data association that is invariant to noise, some approaches explicitly estimate and remove FPN online \cite{polizzi2022data, chen2024eamt, hou2025robust, wu2025monocular}. However, \cite{chen2024eamt} and \cite{hou2025robust} rely on a bilateral filter to isolate FPN, ignoring its spatially smooth component, while \cite{polizzi2022data} and \cite{wu2025monocular} use the method in \cite{das2021online}, which requires pixel correspondences that may be hard to obtain prior to the noise removal (a chicken-and-egg problem, as the authors acknowledge \cite{das2021online}). The method in \cite{das2021online} also estimates FPN only at sparse pixels, interpolating the rest. Our method fits in this latter category, but estimates FPN densely without spatial filtering or correspondences.

It is important to note that thermal cameras also have built-in methods of mitigating FPN. However, shutter-based non-uniformity corrections (NUCs) pose a risk to tracking by pausing the image stream \cite{borges2016practical}, and onboard noise filtering introduces black-box multi-view inconsistency \cite{specarmi_trnerf}. To our knowledge, we are the first to investigate this latter issue in SLAM, revealing that onboard filters (often enabled by default) can be very harmful in low contrast scenes.

Despite the susceptibility of microbolometers to motion blur and rolling shutter distortions, these topics have been rarely covered in the thermal SLAM context. While several thermal SLAM papers \cite{papachristos2018thermal,khattak_robust_2019, khattak2020complementary, doer_radar_2021, flemmen2021rovtio} have applied or built upon ROVIO \cite{bloesch_iterated_2017, bloesch_robust_2015}, a blur-robust visual-inertial odometry method, we believe our method is the first to integrate the unique microbolometer model of motion blur. Moreover, to our knowledge, ours is the first thermal SLAM method to consider rolling shutter distortions. 

\paragraph{NeRF \& 3DGS Restoration}

NeRF \cite{mildenhall_nerf_2020} and 3DGS \cite{kerbl_3d_2023} have proven highly effective methods for novel view synthesis and 3D reconstruction. While each originally assumed sharp, global shutter inputs, it has since been demonstrated that accurate NeRF and 3DGS scenes can be trained despite motion blur \cite{wang_bad-nerf_2023,lee_exblurf_2023,zhao_bad-gaussians_2024, chen_deblur-gs_2024}, rolling shutter distortions \cite{li_usb-nerf_2023, wu20253dgut}, or a combination of these degradations \cite{specarmi_trnerf, seiskari_gaussian_2024}. In particular, TRNeRF restores highly degraded thermal images by modeling the microbolometer in the NeRF rendering pipeline \cite{specarmi_trnerf}. We translate this idea into 3DGS, improving efficiency, and extend the FPN parameterization. More importantly, we eliminate TRNeRF's requirement for ground truth poses by integrating these concepts into a SLAM system.

\paragraph{NeRF \& 3DGS SLAM}

Initially parallel to the work on restoration, NeRF \cite{mildenhall_nerf_2020} and 3DGS \cite{kerbl_3d_2023} have been increasingly used as dense, high fidelity map representations in SLAM \cite{tosi_how_2024}. NeRF and 3DGS SLAM methods can be broadly categorized as frame-to-frame or frame-to-model \cite{tosi_how_2024}. Frame-to-frame methods typically employ existing SLAM methods (e.g., ORB-SLAM3 \cite{campos_orb-slam3_2021} or DROID-SLAM \cite{teed_droid-slam_2021}) as frontends, optimizing the NeRF or 3DGS map in the backend. Frame-to-model methods utilize the NeRF or 3DGS map across the entire system, performing tracking directly against the map. As our benchmark suggests (\cref{section:exp_slam}), existing frame-to-frame tracking methods struggle with the degradations we target, pointing to the frame-to-model strategy as more promising. Monocular(-inertial) frame-to-model approaches are much fewer in number. Among them, MonoGS \cite{Matsuki:Murai:etal:CVPR2024} stands out from other methods \cite{zhu2024nicer, sun2024mm3dgs, mansour2025udgs} that rely on monocular depth networks (unreliable with degraded thermal images). We therefore draw the most inspiration from MonoGS \cite{Matsuki:Murai:etal:CVPR2024}, especially in map initialization, keyframe selection, and Gaussian creation, while also substantially expanding upon it and developing our system from scratch around a different rasterizer (gsplat \cite{ye2025gsplat}). 

Recently, the lines of research on restoration and SLAM have intersected in multiple papers on blur-aware NeRF- and 3DGS-based visual SLAM \cite{bae_i2-slam_2024,wang_mba-slam_2024,girlanda2025deblur}. However, none fuse with an IMU or utilize a B-spline trajectory, and the monocular demonstrations rely on frame-to-frame tracking \cite{bae_i2-slam_2024, girlanda2025deblur} and monocular depth estimates \cite{girlanda2025deblur}. There are also several NeRF and 3DGS SLAM papers that share elements with our system, but do not leverage them to the same extent. TS-SLAM \cite{he2025optimizing} is an RGB-D method that jointly optimizes a B-spline trajectory together with a NeRF map, but uses a dynamics prior rather than fusing with an IMU. \cite{wu2025monocular} proposes a thermal monocular NeRF SLAM method that, like ours, estimates FPN, but does so through pixel correspondences (as noted above). Moreover, it relies on a visual SLAM frontend, DSM \cite{zubizarreta2020direct}, that is not blur or rolling shutter-aware. Finally, existing monocular-inertial methods limit the direct influence of the IMU to their frontends \cite{vingsmono, liao2024vi, bai2024real} or, more narrowly, to initializing the camera pose in tracking \cite{sun2024mm3dgs}. In contrast, we use an IMU loss directly in optimizing the 3DGS map, pivotally including gyroscope data from the beginning.

\paragraph{B-Splines in Rolling Shutter SLAM}

We draw inspiration from many rolling shutter-aware SLAM methods that have demonstrated the utility of continuous B-spline trajectories \cite{lovegrove_spline_2013,patron-perez_spline-based_2015, kerl2015dense, kim2016direct, lang_ctrl-vio_2022}. As originally observed in \cite{lovegrove_spline_2013}, this representation naturally supports estimation of the range of poses contributing to each rolling shutter frame, while also enabling residuals to be computed directly against IMU measurements (without need for IMU (pre-)integration). To support our SLAM method and future work, we contribute a stand-alone package for efficient, on-manifold, B-spline trajectory optimization with Pytorch (detailed in \cref{section:bpline_impl_details}\footnote{Supplementary material elements are referred to only by their alphanumeric section labels for brevity}). 
\section{Method}

\begin{figure}
\centering
\includegraphics[width=1.0\columnwidth]{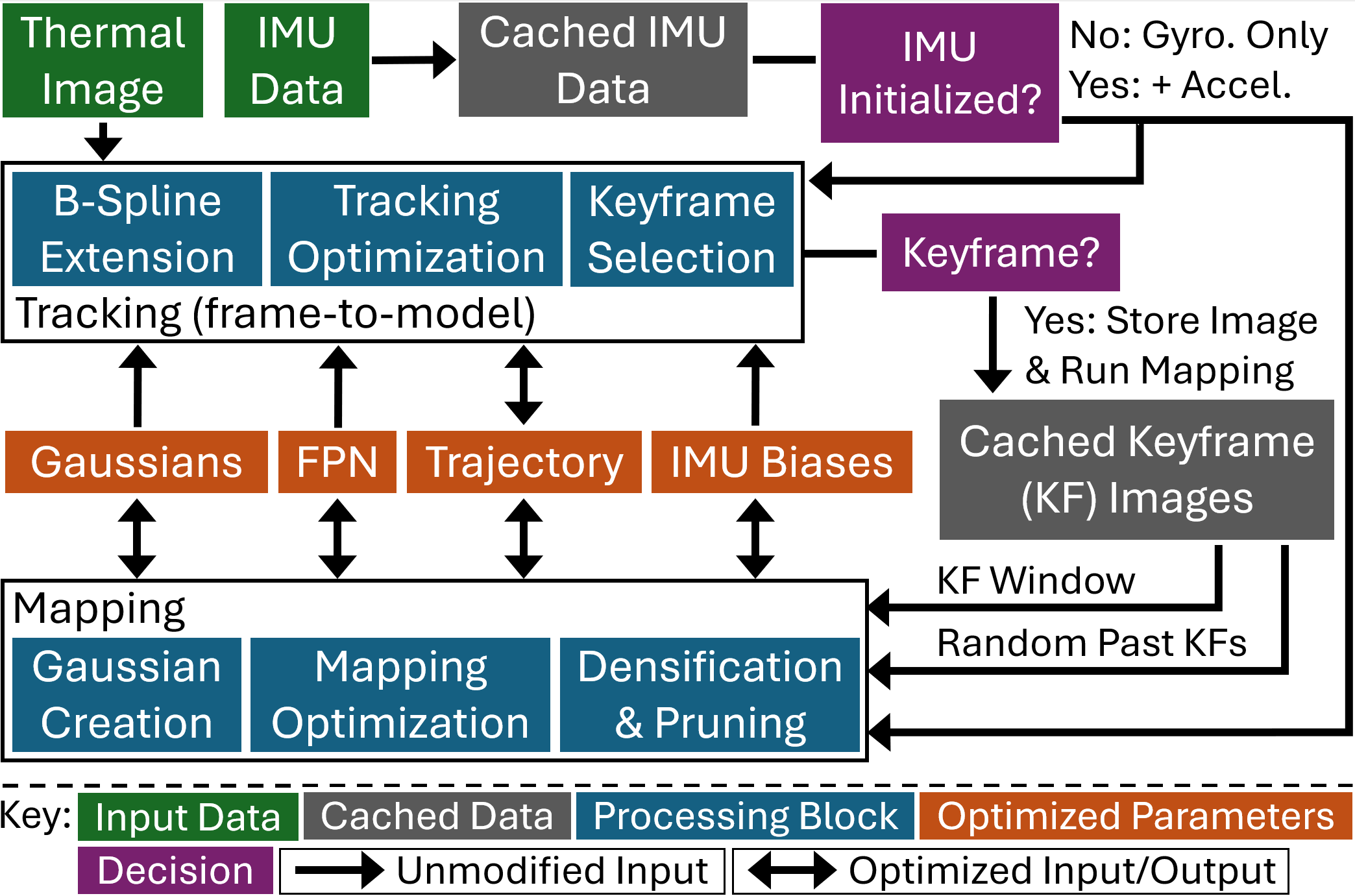}
\caption{TRGS-SLAM system diagram\vspace{-4mm}}  
\label{figure:flowchart}
\end{figure}

\Cref{figure:flowchart} presents a high level overview of TRGS-SLAM, highlighting the data and parameters flowing in and out of tracking (\cref{section:tracking}) and mapping (\cref{section:mapping}). The B-spline and IMU loss (\cref{section:bspline_traj_and_imu_loss}) are used throughout the system, as are the 3DGS map and microbolometer-aware rendering method (\cref{section:deg_aware_rendering}). These components involve the learned IMU biases and FPN, respectively.

\subsection{Preliminary: Microbolometer Image Model}
\label{section:microbolo_model}

In a raw microbolometer thermal image, the value of pixel $(x_p, y_p)$ at time $t$ can be given by \cite{specarmi_trnerf}:
\begin{align}
    \tilde{n}'_{x_p, y_p}(t) &= \tilde{m}'_{x_p, y_p}(t) + \tilde{o}'_{x_p, y_p}(t) \label{equation:final_pixel} \\
    \tilde{m}'_{x_p, y_p}(t) &= \frac{1}{\tau}\int_{-\infty}^{t} \exp\left({\frac{s-t}{\tau}}\right) \tilde{p}'_{x_p, y_p}(s)ds \label{equation:just_blur}
\end{align}
\Cref{equation:just_blur} describes the distinct model of motion blur in microbolometers, producing blurry pixel value $\tilde{m}'_{x_p, y_p}(t)$. $\tilde{p}'_{x_p, y_p}(t)$ is a value directly proportional to the power incident on pixel $(x_p, y_p)$ at time $t$ and $\tau$ is the camera's thermal time constant. \Cref{equation:final_pixel} gives the final pixel value $\tilde{n}'_{x_p, y_p}(t)$, modeling slowly varying FPN as a pixelwise offset $\tilde{o}'_{x_p, y_p}(t)$. The prime symbol denotes that these are quantities relating to a lens-distorted image and the tilde denotes that these are measured quantities. Additionally, microbolometers employ a rolling shutter readout such that the value of pixel $(x_p, y_p)$ in image $l$ is obtained by evaluating \cref{equation:final_pixel} at time $t'_{x_p, y_p, l}$ \cite{specarmi_trnerf}:
\begin{equation}
    t'_{x_p, y_p, l} = t'_{0, 0, l} + x_p\Delta t_{\text{pix}} + y_p w \Delta t_{\text{pix}} 
\label{equation:rs_readout}
\end{equation}
where $t'_{0, 0, l}$ is the time the top left pixel $(0, 0)$ is read out, $\Delta t_{\text{pix}}$ is the readout delay between each pixel, and $w$ is the width of the image. We refer readers to \cite{specarmi_trnerf} for more detail.

\subsection{B-Spline Trajectory \& IMU Loss}
\label{section:bspline_traj_and_imu_loss}

\begin{figure*}
\centering
\includegraphics[width=1.0\textwidth]{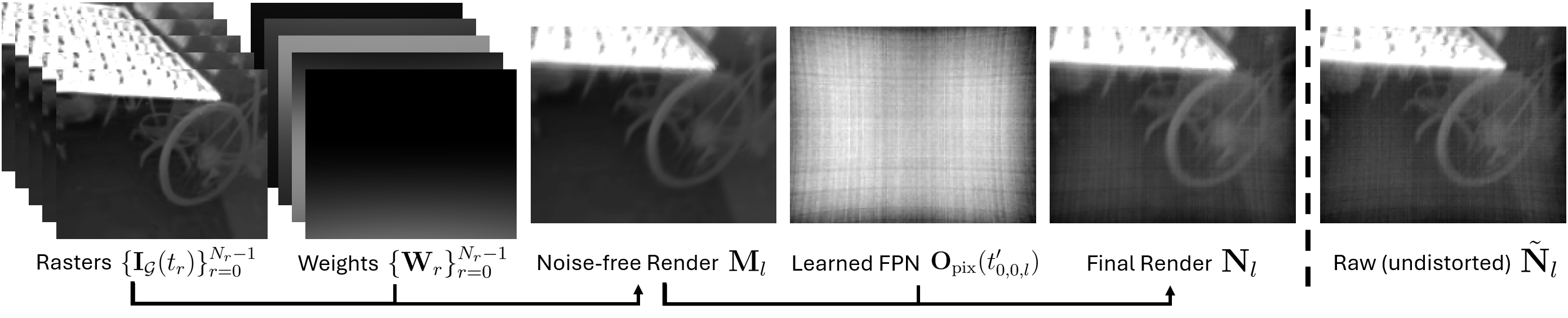}
\caption{Microbolometer-aware rendering process. A sum of multiple rasters, each multiplied with precomputed pixel-wise weights, is combined with a learned FPN estimate to obtain an approximation of the degraded input image.\vspace{-4mm}}  
\label{figure:aware_rendering}
\end{figure*}

We represent the continuous-time trajectory with two uniform B-splines, one for position $\mathbf{p}(t)\in\mathbb{R}^3$ and one for rotation $\mathbf{R}(t)\in\text{SO}(3)$ (as defined in \cref{section:bpline_formulation}). The B-splines' values are determined by control points, $\mathbf{p}_i$ and $\mathbf{R}_i$, knots, $t_i$, and the spline order, $k$. The value of a B-spline at any time $t$ is influenced by $k$ active control points and the trajectory is estimated through their optimization.  

The relative scale trajectory of the camera in the world frame will be denoted as $\mathbf{\bar{T}}_{\text{C}}^{\text{W}}(t)$, where the bar indicates the relative scale. Prior to IMU initialization  (\cref{section:imu_initialization}), the position and rotation splines represent this trajectory directly: $\mathbf{\bar{T}}_{\text{C}}^{\text{W}}(t)=[\mathbf{R}_{\text{C}}^{\text{W}}(t), \mathbf{\bar{p}}_{\text{C}}^{\text{W}}(t)]$. After IMU initialization, the position and rotation splines represent the absolute scale trajectory of the IMU, $\mathbf{T}_{\text{I}}^{\text{W}}(t)=[\mathbf{R}_{\text{I}}^{\text{W}}(t), \mathbf{p}_{\text{I}}^{\text{W}}(t)]$, and $\mathbf{\bar{T}}_{\text{C}}^{\text{W}}(t)$ is evaluated using the calibrated IMU extrinsics $\mathbf{T}_{\text{C}}^{\text{I}}(t)=[\mathbf{R}_{\text{C}}^{\text{I}}, \mathbf{p}_{\text{C}}^{\text{I}}]$ and estimated scale factor $S$:
\begin{equation}
\mathbf{\bar{T}}_{\text{C}}^{\text{W}}(t)=[\ \mathbf{R}_{\text{I}}^{\text{W}}(t)\mathbf{R}_C^I \ , \ \frac{1}{S} \left(\mathbf{R}_{\text{I}}^{\text{W}}(t) \mathbf{p}_{\text{C}}^{\text{I}}  + \mathbf{p}_{\text{I}}^{\text{W}}(t)\right) \ ]
\label{equation:abs_imu_to_rel_cam}
\end{equation}

B-splines have closed form derivatives and are $\mathbf{C}^{k-2}$ continuous (see \cref{section:bpline_derivatives} for details). Therefore, $k>=4$ ensures continuous acceleration and enables direct comparisons between derivatives of the estimated trajectory and IMU measurements. Prior to IMU initialization, we compute a residual $\mathbf{r}_{g, m}$ against gyroscope measurement $\tilde{\boldsymbol{\omega}}_{\text{I}, m}^{\text{I}}$ taken at time $t_m$ as:
\begin{equation}
\mathbf{r}_{g, m} = \mathbf{R}_{\text{C}}^{\text{I}}\boldsymbol{\omega}_{\text{C}}^{\text{C}}(t_m) - (\tilde{\boldsymbol{\omega}}_{\text{I}, m}^{\text{I}} - \mathbf{b}_{g}) 
\end{equation}
where $\boldsymbol{\omega}_{\text{C}}^{\text{C}}(t)$ is the estimated angular velocity and $\mathbf{b}_{g}$ is the gyroscope bias. After IMU initialization, we compute a gyroscope $\mathbf{r}_{g, m}$ and accelerometer $\mathbf{r}_{a, m}$ residual as:
\begin{equation}
\begin{gathered}
\mathbf{r}_{g, m} = \boldsymbol{\omega}_{\text{I}}^{\text{I}}(t_m) - (\tilde{\boldsymbol{\omega}}_{\text{I}, m}^{\text{I}} - \mathbf{b}_{g}) \\
\mathbf{r}_{a, m} = \mathbf{a}_{\text{I}}^{\text{W}}(t_m) - \left(\mathbf{R}_{\text{I}}^{\text{W}}(t_m)(\tilde{\mathbf{a}}_{\text{I}, m}^{\text{I}} - \mathbf{b}_{a}) + G \hat{\mathbf{g}}^{\text{W}}\right)
\end{gathered}
\end{equation}
where $\boldsymbol{\omega}_{\text{I}}^{\text{I}}(t)$, $\mathbf{a}_{\text{I}}^{\text{W}}(t)$, and $\mathbf{R}_{\text{I}}^{\text{W}}(t)$ are the estimated angular velocity, linear acceleration, and rotation, $\tilde{\mathbf{a}}_{\text{I}, m}^{\text{I}}$ is an accelerometer measurement taken at time $t_m$, $\mathbf{b}_{a}$ is the accelerometer bias, $\hat{\mathbf{g}}^{\text{W}}$ is the direction of gravity in the world frame (as a unit vector), and $G$ is the magnitude of gravity. Note that $\mathbf{b}_{g}$ and $\mathbf{b}_{a}$ are learned parameters, while $\hat{\mathbf{g}}^{\text{W}}$ is held constant outside of periodic updates (\cref{section:imu_initialization}).

Given a set of IMU measurements $\mathcal{I}$ we compute an IMU-based loss $\mathcal{L}_{\text{IMU}, \mathcal{I}}$ as follows:
\begin{equation}
\mathcal{L}_{\text{IMU}, \mathcal{I}} = \sum_{m\in\mathcal{I}} \lambda_{g}\left\| \mathbf{r}_{g, m} \right\|_2 + \lambda_{a}\left\| \mathbf{r}_{a, m} \right\|_2
\label{equation:imu_loss}
\end{equation}
where $\lambda_{\text{g}}$ and $\lambda_{\text{a}}$ are scalar coefficients. Note that the terms involving the accelerometer are excluded prior to IMU initialization. Additionally, we compute a loss from the learned biases using another set coefficient $\lambda_{\text{bias}}$:
\begin{equation}
\mathcal{L}_{\text{bias}} = \lambda_{\text{bias}}\left(\left\| \mathbf{b}_{g} \right\|_2 + \left\| \mathbf{b}_{a} \right\|_2\right)
\end{equation}
to prevent overfitting with outlier residuals.  
 
\subsection{3DGS Map \& Microbolometer-Aware \\ Rendering}
\label{section:deg_aware_rendering}

We represent the map as a set of 3D Gaussians $\mathcal{G}$, each defined by a scalar opacity $\sigma_g$ and intensity $c_g$, and a relative scale 3D covariance ellipsoid parameterized by a mean $\bar{\boldsymbol{\mu}}_g$, scale $\bar{\mathbf{s}}_g$, and quaternion $\mathbf{q}_g$ in the world frame. In the original 3DGS method \cite{kerbl_3d_2023}, spherical harmonics are used to represent view-dependent color. Here, we assume no view dependency to reduce the number of learnable parameters, and we require only a scalar intensity as thermal images are single-channel. 

In 3DGS \cite{kerbl_3d_2023}, images are rasterized by splatting (projecting) the 3D Gaussians onto the 2D image plane and computing a pixel value $p$ as:
\begin{equation}
    p = \sum_{g\in\mathcal{B}} c_g \alpha_g \prod_{j=1}^{g-1} (1-\alpha_j)
\label{equation:3dgs_raster_eq}
\end{equation}
where $\mathcal{B}$ is a depth-ordered subset of Gaussians visible to the pixel and $\alpha_g$ is obtained by multiplying the Gaussian's opacity $\sigma_g$ by the Gaussian's 2D distribution evaluated at the pixel coordinate. The primary inputs to the 3DGS method are a set of sharp, global shutter images $\{\tilde{\mathbf{I}}'_l\}_{l=0}^{N_\text{img}-1}$ and corresponding camera pose estimates $\{\bar{\mathbf{T}}_{\text{C}, l}^{\text{W}}\}_{l=0}^{N_\text{img}-1}$. The 3D to 2D projection assumes an ideal pinhole camera model, so lens undistortion is applied to the input images to obtain $\{\tilde{\mathbf{I}}_l\}_{l=0}^{N_\text{img}-1}$. In training, the loss is computed by rasterizing an image $\mathbf{I}_{\mathcal{G}, l}$ at pose $\bar{\mathbf{T}}_{\text{C}, l}^{\text{W}}$ and directly comparing it to $\tilde{\mathbf{I}}_l$. To adapt this to degraded thermal images, we extend the rendering pipeline to rasterize multiple images, blend them together, and sum the result with learned FPN.

Let $[\tilde{\mathbf{N}}_l]_{y_p, x_p} = \tilde{n}_{x_p, y_p}(t_{x_p, y_p, l})$ denote a raw microbolometer image after downsampling and lens undistortion. We show in \cref{section:undistorted_microbolo_model} that $\tilde{n}_{x_p, y_p}(t_{x_p, y_p, l})$ approximately follows the form of \cref{equation:final_pixel} at an effective readout time $t_{x_p, y_p, l}$. To render an estimate of $\tilde{\mathbf{N}}_l$, we begin by computing $[\mathbf{M}_l]_{y_p, x_p} = m_{x_p, y_p}(t_{x_p, y_p, l})$. To do this, we rasterize $N_r$ images at times $\{t_r\}_{r=0}^{N_r-1}$ evenly spaced over $[t_{\text{max}, l}-T_b, t_{\text{max}, l}]$, where $t_{\text{max}, l}$ is the latest effective readout time across $\tilde{\mathbf{N}}_l$, and $T_b$ is the maximum integration interval. This rasterization involves evaluating the position and rotation splines at each time to obtain $\{\mathbf{\bar{T}}_{\text{C}}^{\text{W}}(t_r)\}_{r=0}^{N_r-1}$. Treating the pixel values of the rasterized images $\{\mathbf{I}_{\mathcal{G}}(t_r)\}_{r=0}^{N_r-1}$ as samples $\{p_{x_p, y_p}(t_r)\}_{r=0}^{N_r-1}$, we can compute $m_{x_p, y_p}(t_{x_p, y_p, l})$ as a discrete integral in the form of \cref{equation:just_blur}. Ultimately, as detailed in \cref{section:discrete_motion_blur}, the computation of $\mathbf{M}_l$ can be efficiently cast as a sum of element-wise products:
\begin{equation}
\mathbf{M}_l = \sum_{r=0}^{N_r-1} \mathbf{I}_{\mathcal{G}}(t_r) \odot \mathbf{W}_r
\end{equation}
where $\{\mathbf{W}_r\}_{r=0}^{N_r-1}$ are precomputed weights that are a function of $N_r$, $T_b$, $\tau$, $\Delta t_{\text{pix}}$, and the lens undistortion maps. 

To model the slowly time varying FPN, we use two splines: $\mathbf{O}_{\text{pix}}(t)\in\mathbb{R}^{h \times w}$ and $o_{\text{global}}(t)\in\mathbb{R}$, where $w$ and $h$ are the image width and height, respectively. $\mathbf{O}_{\text{pix}}(t)$ models pixel-wise FPN and, similar to TRNeRF \cite{specarmi_trnerf}, we use the following loss to encourage $\mathbf{O}_{\text{pix}}(t)$ to be zero-mean:
\begin{equation}
    \mathcal{L}_{\text{FPN}}(t) = \frac{\lambda_{\text{FPN}}}{wh}\left|\sum_{x_p=0}^{w-1}\sum_{y_p=0}^{h-1} [\mathbf{O}_{\text{pix}}(t)]_{y_p, x_p} \right|
\end{equation}
where $\lambda_{\text{FPN}}$ is a set coefficient. As $\mathbf{O}_{\text{pix}}(t)$ is zero-mean, we use $o_{\text{global}}(t)$ to separately account for global (image-wide) intensity changes. Both of these splines are constructed similarly to the position spline (see \cref{section:bpline_formulation}). The control points of $\mathbf{O}_{\text{pix}}(t)$ are stored as vectors in $\mathbb{R}^{wh}$ and the result of evaluating the spline is reshaped. Note that the spline knot intervals can be used to independently control how slowly these FPN components vary.  

Putting these parts together, the final render for image $\mathbf{N}_l$ is computed as follows:
\begin{equation}
\mathbf{N}_l = \mathbf{M}_l + \mathbf{O}_{\text{pix}}(t'_{0, 0, l}) + o_{\text{global}}(t'_{0, 0, l})
\end{equation}
The process is visualized in \cref{figure:aware_rendering}. We define the loss for image $l$ as the mean absolute error between the rendered image and the undistorted image:
\begin{equation}
\mathcal{L}_{\text{img}, l} = \frac{1}{wh}\|\mathbf{N}_l - \tilde{\mathbf{N}}_l\|_{1}
\end{equation}

\subsection{System Initialization}
\label{section:map_initialization}

The position and rotation splines are initialized with all control points set to zero and identity, respectively (i.e., the initial camera frame defines the world frame). All control points of both FPN splines are also initialized to zero. The map is initialized as a plane of $N_{\text{init}}$ Gaussians one unit (unscaled) in front of the camera. This is done by randomly sampling $N_{\text{init}}$ pixels across the first image. The Gaussian means are initialized by back projecting the pixels with a depth of 1 and the pixel values are used to set the intensities. The opacities are set to a default value, the quaternions are randomized, and the scale of each Gaussian is set to the average distance of its three nearest neighbors. The Gaussians are subsequently fit to the first image through $N_{\text{iter}, \text{init}}$ iterations of optimization with the following loss: 
\begin{equation}
    \mathcal{L}_{\text{init}} = \frac{1}{wh}\| \mathbf{I}_{\mathcal{G}}(t'_{0, 0, 0}) - \tilde{\mathbf{N}}_0\|_{1}
\end{equation}
Note that at this stage only the Gaussians are optimized and blur and rolling shutter are not yet modeled. For stability, blur and rolling shutter are not modeled until after a set number of initial keyframes.  

\subsection{Tracking}
\label{section:tracking}
After initialization, tracking is applied to each new image.

\paragraph{B-Spline Extension}\label{section:bspline_ext}
The splines can be evaluated for $t \in [t_0, t_0 + (N_c - k + 1) \Delta t_c)$, where $N_c$ is the total number of control points and $\Delta t_c$ is the knot interval. When a new image $l$ arrives, control points must be added to extend the splines to the end of readout $t_{\text{max}, l}$ prior to tracking optimization. In the early stages of our system, this is done by simply duplicating the last control point of each spline. After a set number of keyframes, we additionally compute $N_q$ predicted poses at times $\{t_q\}_{q=0}^{N_q-1}$ evenly spaced over $[t'_{0, 0, l-1}, t_{\text{max}, l}]$ and fit the active set of control points to these predictions. Prior to IMU initialization, predictions are made assuming constant velocity; after IMU initialization, predictions are made through IMU integration.  Further details on this process are provided in \cref{section:bspline_ext_details}. The FPN splines, $\mathbf{O}_{\text{pix}}(t)$ and $o_{\text{global}}(t)$, must also be extended occasionally. When necessary, we extend the FPN splines by duplicating their last control points, without fitting them to any predictions.

\paragraph{Tracking Optimization}
In tracking we optimize the position and rotation control points only. All other parameters are frozen. Optimization is performed for a maximum of $N_{\text{iter}, \text{track}}$ iterations with the following loss:
\begin{equation}
    \mathcal{L}_{\text{track}} = \frac{1}{\left|\mathcal{I}_{\text{track}}\right|}\mathcal{L}_{\text{IMU}, \mathcal{I}_{\text{track}}} + \mathcal{L}_{\text{img}, l}
\end{equation}
where $\mathcal{I}_{\text{track}}$ is the set of IMU measurements taken over $[\min(t_{0, 0, l-1}, t_{\text{max},l}-T_b), t_{\text{max}, l}]$. Optimization is considered converged if the average change in both the positions and rotations across the raster timestamps falls below set thresholds.  

\paragraph{Keyframe Selection \& Window Management}
After tracking optimization, we follow MonoGS \cite{Matsuki:Murai:etal:CVPR2024} and select the current image as a keyframe if it satisfies one of two conditions: the translation since the last keyframe (normalized by the relative scale median depth) exceeds a threshold, or the intersection over union between the currently visible Gaussians and those visible to the last keyframe falls beneath a threshold. Unlike MonoGS, our keyframe window is defined simply as the last $N_{\text{window}}$ keyframes. We have found that the random keyframe sampling in mapping is sufficient to provide wide-baseline multi-view constraints.

\subsection{Mapping}
\label{section:mapping}
When a new keyframe is selected, mapping is applied.

\paragraph{Gaussian Creation}
Prior to mapping optimization, new Gaussians are created by randomly sampling and back projecting pixels in the new keyframe, as done with the first image in \cref{section:map_initialization}. However, in this case, the depth of new Gaussians is set to the relative scale median depth in the new keyframe, and the sampling is restricted to pixels with an accumulated alpha value beneath a set threshold (where the accumulated alpha value is given by \cref{equation:3dgs_raster_eq} with $c_g$ removed). The number of new Gaussians is set as $N_{\text{max}} \ N_{\text{cand}} / (wh)$ where $N_{\text{max}}$ is a set parameter and $N_{\text{cand}}$ is the number of candidate pixels (as determined by the alpha threshold). This avoids creating new Gaussians in previously mapped areas. 

\paragraph{Mapping Optimization}
In mapping we jointly optimize all learned parameters: the IMU biases, the Gaussians, and the position, rotation, and FPN control points. Optimization is performed for $N_{\text{iter}, \text{map}}$ iterations with the following loss:
\begin{equation}
\begin{aligned}
    &\mathcal{L}_{\text{map}} = 
    \frac{1}{|\mathcal{I}_{\text{map}}|} \left(\lambda_w \mathcal{L}_{\text{IMU}, \mathcal{W_I}}+\lambda_s \mathcal{L}_{\text{IMU}, \mathcal{S_I}}\right) + \mathcal{L}_{\text{bias}} + \\
    & \frac{1}{|\mathcal{K}|}\left(\lambda_w \sum_{l\in\mathcal{W}_K} \mathcal{L}_{\text{img}, l}+\lambda_s \sum_{l\in\mathcal{S}_K} \mathcal{L}_{\text{img}, l}+\sum_{l\in\mathcal{K}} \mathcal{L}_{\text{FPN}}(t'_{0, 0, l})\right)
\end{aligned}
\label{equation:mapping_loss}
\end{equation}
where $\mathcal{W}_K$ is the set of all keyframes in the current window, $\mathcal{S}_K$ is a set of past keyframes randomly sampled at each iteration, and $\mathcal{K}=\mathcal{W}_K\cup\mathcal{S}_K$. $\mathcal{W}_I$ and $\mathcal{S}_I$ are disjoint sets of IMU measurements associated with $\mathcal{W}_K$ and $\mathcal{S}_K$, and $\mathcal{I}_{\text{map}}=\mathcal{W}_I\cup\mathcal{S}_I$. The coefficients $\lambda_w$ and $\lambda_s$ balance the influence of the keyframe window relative to the past keyframes. Without these coefficients, we observed that pose drift was not consistently corrected. Rather, the keyframe window frequently overwrote previously observed portions of the map. We address this by setting $\lambda_s>\lambda_w$ to give more weight to the past keyframes. The coefficients are applied to both the image and the IMU losses to maintain balance between them. Additionally, we manually modify the gradients of the trajectory control points and Gaussians to compensate for this weighting scheme. The full details are given in \cref{section:mapping_opt_details}.

\begin{figure}
\centering
\includegraphics[width=1.0\columnwidth]{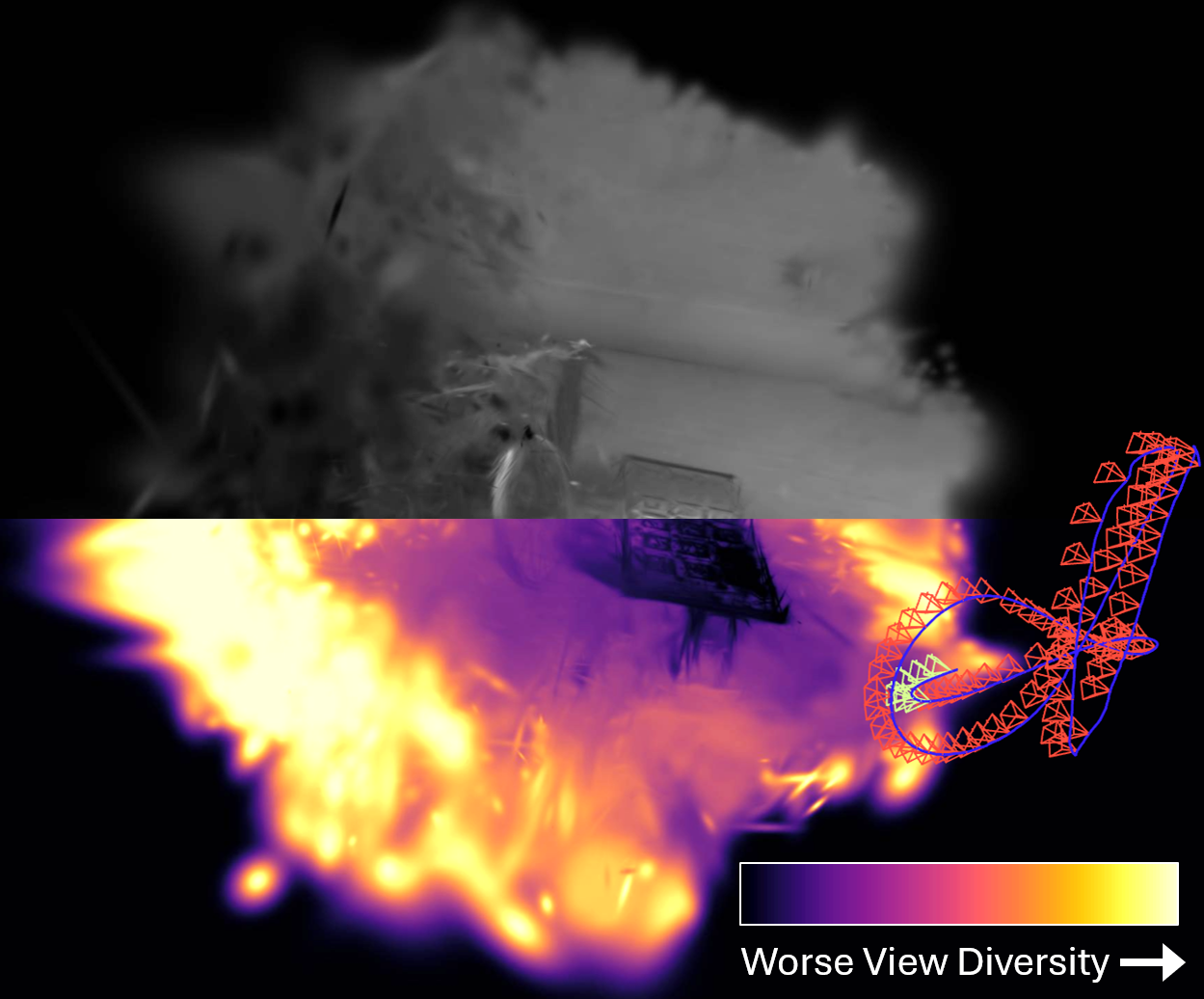}
\caption{A depiction of the Gaussians targeted by our view-diversity-based opacity resetting. The color scale is mapped to the log of the condition number of $\mathbf{V}_g$ for each Gaussian.\vspace{-4mm}}  
\label{figure:view_diversity}
\end{figure}

\paragraph{View-Diversity-Based Opacity Resetting}
The original 3DGS paper \cite{kerbl_3d_2023} introduces several rules that are periodically applied to split, duplicate, prune, and reset Gaussians. These rules serve to populate empty areas and reconstruct detail. One such rule, intended to reduce the number of Gaussians, is to periodically reset the Gaussian opacities to a low value. In combination with a rule to prune low opacity Gaussians, this allows subsequent optimization to recover necessary Gaussians only and cull the rest. However, this global resetting is inefficient and destabilizing in the SLAM context. Therefore, we limit this resetting to Gaussians with poor view diversity. Specifically, at each mapping iteration we compute the following for each Gaussian $g$:
\begin{equation}
    \mathbf{V}_g = \mathbf{V}_{g, \text{prev}} + \sum_{r\in\mathcal{R}_g} \hat{\mathbf{d}}_{r, g} \hat{\mathbf{d}}_{r, g}^T
\end{equation}
where $\mathbf{V}_{g, \text{prev}}$ is the result from the previous iteration, $\mathcal{R}_g$ is the subset of rasters the Gaussian appears in, and $\hat{\mathbf{d}}_{r, g}$ is the unit vector pointing from the camera center at raster $r$ to the Gaussian mean $\bar{\boldsymbol{\mu}}_g$. $\mathbf{V}_g$ is therefore the second moment matrix of these view directions, accumulated over many mapping iterations. When it comes time to reset opacities, we compute the condition number of each $\mathbf{V}_g$ and reset Gaussians with a condition number above a set threshold. $\mathbf{V}_g$ is then reset to $\mathbf{0}$. As shown in \cref{figure:view_diversity}, this tends to eliminate poorly constrained Gaussians at the edges of the scene.

We maintain the remaining rules for splitting, duplicating, and pruning. However, as detailed in \cref{section:modified_gradient_accum}, the gradient accumulation involved in these rules must be modified to accommodate our microbolometer-aware rendering. When the rules are applied, the mapping iteration counter is reset to ensure the map is stable before tracking resumes.

\paragraph{Relocalization} When the current keyframe image $l$ observes a new area, mapping is largely unconstrained and consistently results in a high SSIM value between the noise-free render $\mathbf{M}_l$ and denoised input image $\tilde{\mathbf{N}}_l - \mathbf{O}_{\text{pix}}(t'_{0, 0, l}) - o_{\text{global}}(t'_{0, 0, l})$. The SSIM is also high when the camera is accurately localized in a pre-existing portion of the map. However, given that the weighting scheme in \cref{equation:mapping_loss} prevents overwriting, mapping can fail to fit the Gaussians to the current keyframe image when a part of the scene is re-observed after significant drift. Therefore, low SSIM signals tracking failure. We implement a simple scheme to recover from tracking failure. When the SSIM is below a set threshold after mapping, we revert the full system state to before mapping and resume tracking. This effectively pauses mapping until tracking is able to relocalize. During this phase, we reduce the IMU residual coefficients, $\lambda_g$ and $\lambda_a$, to allow pose optimization to temporarily ignore dynamics and jump to the correct solution. 

\subsection{IMU Initialization \& Updates}
\label{section:imu_initialization}

IMU initialization solves for the unknown gravity direction $\hat{\mathbf{g}}$ and scale factor $S$ needed to compute a loss against the accelerometer. It also produces an estimate of the gyroscope and accelerometer biases, $\mathbf{b}_g$ and  $\mathbf{b}_a$. After a set period, we perform IMU initialization as described in \cite{analytical_imu_init} using keyframe poses sampled at a set rate and all IMU measurements. Using the estimated scale factor, we transform the trajectory to the IMU frame with absolute scale. To smooth the trajectory, we run mapping optimization for set number of iterations with two changes: we use all IMU measurements at each iteration and all keyframes are randomly sampled. As the initial scale estimate can be poor, we periodically rerun the method in \cite{analytical_imu_init} to update the scale and gravity direction. Full details on the application of scale updates are provided in \cref{section:imu_init_and_update_details}.

\subsection{Offline Refinement}
\label{section:offline_refinement}

We perform offline refinement of the SLAM results by running a set number of additional mapping iterations using all IMU data at each iteration and randomly sampling from all frames (not just keyframes). In this process, we operate on the full resolution images, increase the number of rasters per render, introduce the gravity direction unit vector as an additional learned parameter, drop the IMU bias loss, and re-introduce global opacity resetting. 
\section{Experiments}

\subsection{Dataset}
\label{section:exp_dataset}
We evaluate our method's SLAM and restoration performance on the TRNeRF dataset \cite{specarmi_trnerf}. This dataset includes six sequences combining three camera speeds (slow, medium, and fast) with two scenes (indoor and outdoor). We refer to these sequences by the first letter of the speed and scene (e.g., SO denotes slow outdoor). These combinations allow for an isolation of variables: blur and rolling shutter are insignificant in the slow sequences, and FPN is most prominent in the low contrast indoor scene. See \cref{section:dataset_details} for further details.

\subsection{Implementation}
\label{section:exp_impl}

We build TRGS-SLAM in PyTorch. For simplicity, the system uses a single process (tracking waits for mapping). We use the gsplat 3DGS library \cite{ye2025gsplat} to benefit from its efficient batched rasterization, camera pose gradients, and implementation of anti-aliasing \cite{yu2024mip}. We reimplement the IMU initialization method of \cite{analytical_imu_init} in Python. We introduce a standalone package for B-spline optimization that uses the algorithms introduced in \cite{sommer_efficient_2020}, and the software libraries PyPose \cite{wang2023pypose} and LieTorch \cite{teed2021tangent}, to achieve efficient and stable on-manifold optimization (detailed in \cref{section:bpline_impl_details}). As performed in \cite{specarmi_trnerf}, we rescale the 16-bit thermal images into the $[0, 1]$ range using predefined thresholds. Here, we use the 0.5 and 99.5 percentiles of the pixel values observed in each sequence. We otherwise run TRGS-SLAM with the same parameters across all sequences, detailed fully in \cref{section:full_parameter_values}.

\subsection{SLAM Evaluation}
\label{section:exp_slam}

\begin{figure}
\centering
\includegraphics[width=1.0\columnwidth]{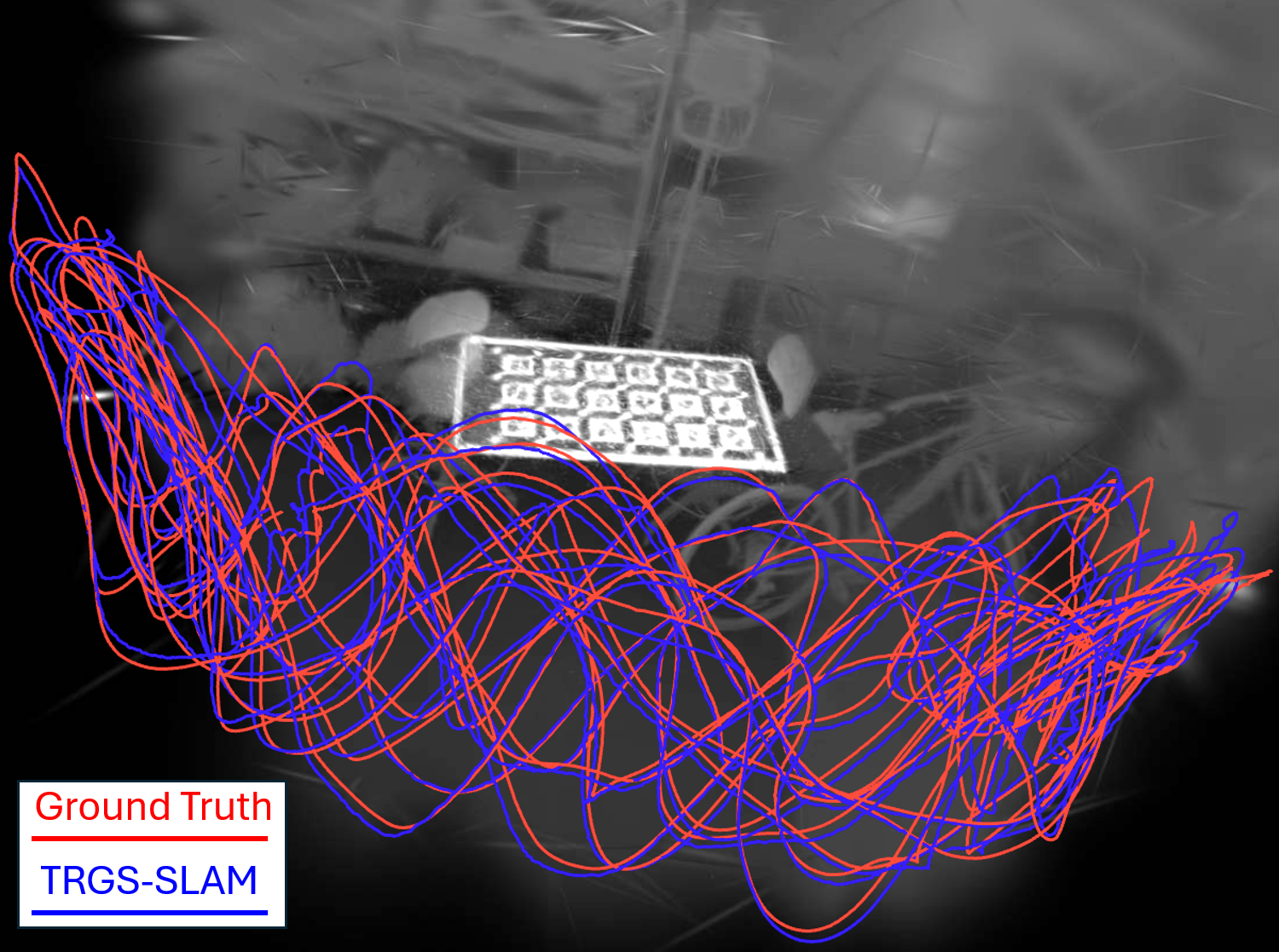}
\caption{TRGS-SLAM result on the medium indoor sequence.\vspace{-2mm}}  
\label{figure:slam_result}
\end{figure}

\begin{table}[]
\centering
\caption{RMSE ATE in cm on the TRNeRF dataset.}
\label{table:slam_results}
\resizebox{\columnwidth}{!}{%
\begin{tabular}{llcccccc}
\hline
\textbf{Type}   & \textbf{Method}          & \textbf{SO}         & \textbf{MO}           & \textbf{FO}           & \textbf{SI}           & \textbf{MI}           & \textbf{FI}           \\ \hline
\multirow{5}{*}{\rotatebox[origin=c]{90}{Monocular}}     & ORB-SLAM3            & 0.8          & 9.4          & $\times$     & $\times$     & $\times$     & $\times$     \\
                                                         & DSM                  & 9.4          & $\times$     & $\times$     & 76.7         & $\times$     & $\times$     \\
                                                         & DROID-SLAM           & \textbf{0.5} & \textbf{2.6} & $\times$     & \textbf{0.7} & $\times$     & $\times$     \\
                                                         & MASt3R-SLAM*         & 1.6          & 94.5         & $\times$     & 2.7          & 88.4         & $\times$     \\
                                                         & MonoGS               & 7.7          & $\times$     & $\times$     & 83.9         & $\times$     & $\times$     \\ \hline
\multirow{5}{*}{\rotatebox[origin=c]{90}{Mono-Inertial}} & ORB-SLAM3 (inertial) & 1.1          & $\times$     & $\times$     & $\times$     & $\times$     & $\times$     \\
                                                         & DM-VIO               & 1.5          & 44.2         & $\times$     & 25.2         & $\times$     & $\times$     \\
                                                         & DBA-VIO              & 1.4          & 25.2         & $\times$     & 2.5          & 26.5         & $\times$     \\
                                                         & ROTIO                & 18.5         & 53.2         & $\times$     & 37.1         & $\times$     & $\times$     \\
                                                         & TRGS-SLAM            & 12.7         & 5.5          & \textbf{2.4} & 6.4          & \textbf{3.9} & \textbf{4.2} \\ \hline
\multicolumn{8}{l}{* RMSE ATE and alignment performed with keyframes only.}  \\
\multicolumn{8}{l}{$\times$ program exited early or error exceeded 1 meter.\vspace{-4mm}}  
\end{tabular}%
}
\end{table}

\begin{figure*}
\centering
\includegraphics[width=1.0\textwidth]{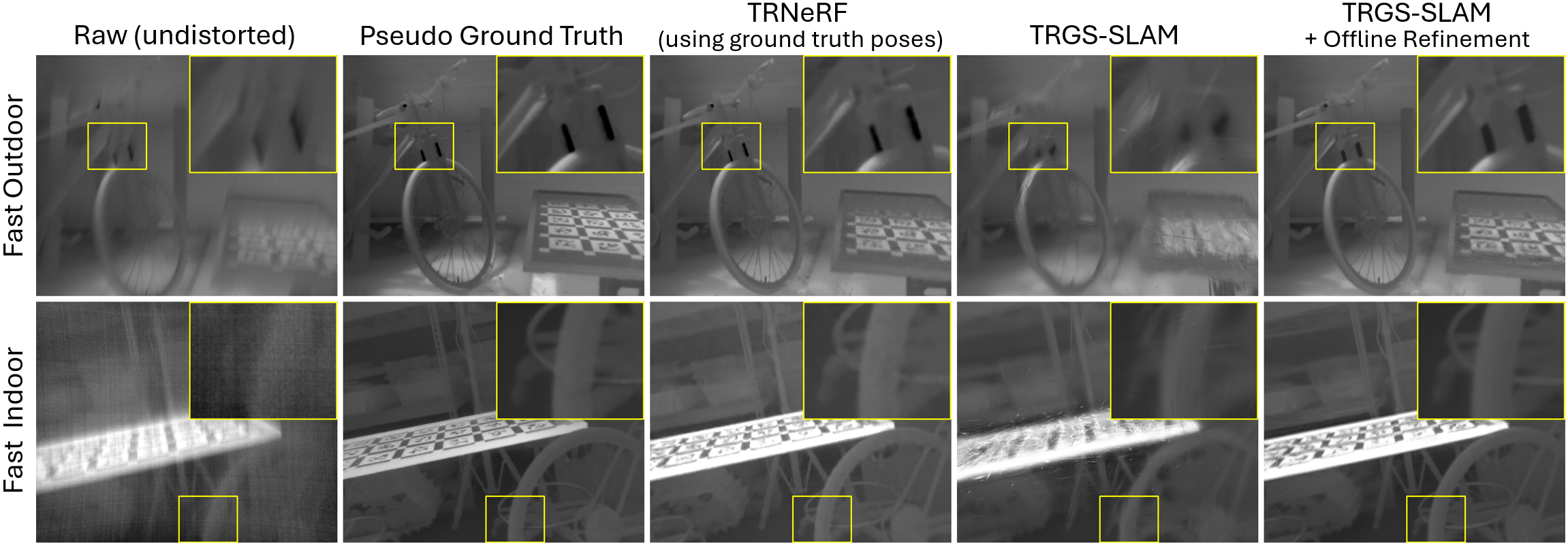}
\caption{Qualitative restoration results. With offline refinement, TRGS-SLAM achieves restoration quality near that of TRNeRF.}  
\label{figure:restoration}
\end{figure*}

We compare TRGS-SLAM against a variety of visual and thermal SLAM methods. To make the case for the frame-to-model paradigm, we test several methods used as frontends in frame-to-frame systems: ORB-SLAM3 \cite{campos_orb-slam3_2021}, DROID-SLAM \cite{teed_droid-slam_2021}, DSM \cite{zubizarreta2020direct}, and DBA-VIO \cite{zhou2024dba} (an inertial extension of DROID-SLAM). We also include the frame-to-model system MonoGS \cite{Matsuki:Murai:etal:CVPR2024}, the direct visual inertial method DM-VIO \cite{stumberg_dm-vio_2022}, and the paradigm-shifting MASt3R-SLAM \cite{murai2025mast3r} (based on two-view reconstruction priors). Although few thermal SLAM methods are publicly available, we are able to test ROTIO \cite{khattak_robust_2019} (using the implementation of \cite{flemmen2021rovtio}), a 16-bit adaptation of ROVIO \cite{bloesch_iterated_2017, bloesch_robust_2015}. With the exception of ROTIO, we rescale the input images with the same thresholds used in our method (and convert to 8-bit), but run on the full resolution images (barring standard resizing procedures \cite{teed_droid-slam_2021, zhou2024dba, murai2025mast3r}). For each sequence, we report the average RMSE ATE across three trials, using $\text{Sim}(3)$ alignment in the pure monocular case and $\text{SE}(3)$ alignment otherwise (using \cite{grupp2017evo}). Unless otherwise noted, the alignment and RMSE ATE calculation are performed using all frames.

\Cref{table:slam_results} gives the results. In the SO sequence, where degradation is nearly absent, TRGS-SLAM is outperformed by multiple baselines. In part, this is due to scale error (our result is 4.7 cm with $\text{Sim}(3)$ alignment). The slow motion makes IMU initialization difficult, though the performance of other inertial methods suggests this aspect of our system could be improved. The value of our method is apparent in the remaining sequences. While some baselines demonstrate impressive robustness under moderate degradation (MO and SI), ours is the only method capable of accurate tracking in the most challenging sequences (FO, MI, and FI). Our result on the MI sequence is shown in \cref{figure:slam_result}.

A thorough ablation study is presented in \cref{section:ablation_details}. The results confirm that to obtain good performance on all sequences it is necessary to use gyroscope data immediately, model all degradations, employ our pose drift corrections schemes, and perform view-diversity-based opacity resetting. Additionally, it is shown that onboard camera noise filters can severely harm performance in the indoor scene.

\subsection{Restoration Evaluation}
\label{section:exp_restore}

\begin{table}[]
\centering
\caption{LPIPS results on the TRNeRF dataset (lower is better) }
\label{table:restoration}
\resizebox{0.9\columnwidth}{!}{%
\begin{tabular}{@{}lcccc@{}}
\toprule
\textbf{Method}                 & \textbf{MO}             & \textbf{FO}             & \textbf{MI}             & \textbf{FI}             \\ \midrule
GS on the Move*        & 0.093          & 0.283          & 0.206          & 0.328          \\
TRNeRF*                 & \textbf{0.042} & \textbf{0.078} & \textbf{0.068} & 0.117          \\
TRGS-SLAM + Refinement & 0.067          & 0.098          & 0.095          & \textbf{0.116} \\ \bottomrule
\multicolumn{5}{l}{*Utilize ground truth poses. Results copied from \cite{specarmi_trnerf}.\vspace{-4mm}}
\end{tabular}%
}
\end{table}

We compare the restoration performance of TRGS-SLAM in combination with offline refinement (\cref{section:offline_refinement}) against TRNeRF \cite{specarmi_trnerf} and the best performing baseline from that paper, GS on the Move \cite{seiskari_gaussian_2024} (both using ground truth poses). The parameters used for offline refinement are given in \cref{section:full_parameter_values}. The LPIPS results are given in \cref{table:restoration} and qualitative results are shown in \cref{figure:restoration}. Remarkably, the results are competitive with TRNeRF, even marginally outperforming it on the FI sequence.

\subsection{Timing \& Memory Analysis}

With a NVIDIA RTX A6000 GPU and an AMD Ryzen 9 5950X CPU, tracking runs at 5-7 Hz and mapping takes 1.3-1.6 seconds per keyframe. Due to the variable frequency of keyframe generation, this leads to an overall FPS of 1-4, with the lowest FPS in the fast sequences. While similar to MonoGS \cite{Matsuki:Murai:etal:CVPR2024}, this FPS falls well short of real time operation with fast motion. More timing information and a broader discussion of limitations are provided in \cref{section:extended_timing} and \cref{section:limits}, respectively. Owing to our Gaussians' scalar intensities (rather than RGB or spherical harmonic values) and our opacity resetting strategy (which limits the total number of Gaussians to $\sim$20k-40k), our system consumes a low amount of memory, peaking at 423 MiB of allocated GPU memory across all SLAM runs.

\section{Conclusion}
\label{section:trgs_conclusions}

We introduce a thermal inertial SLAM system uniquely capable of accurate tracking on severely degraded images and of supporting, through offline refinement, high quality image restoration without known poses. This robustness is achieved by combining a 3DGS map and microbolometer-aware rendering method with a continuous B-spline trajectory and tight IMU fusion. One future direction is the extension to large-scale scenes through degradation aware place recognition and loop closure. Future work could also assess the long-term capability of FPN estimation, and whether shutter-based NUCs can be avoided. Additionally, our method could be applied to rolling shutter visible spectrum cameras, with many of the core elements retained. 
{
    \small
    \bibliographystyle{ieeenat_fullname}
    \bibliography{references}
}

\clearpage
\setcounter{page}{1}
\maketitlesupplementary

\setcounter{section}{0}

\renewcommand\thesection{\Alph{section}}
\renewcommand\thesubsection{\thesection.\arabic{subsection}}

\renewcommand{\theHsection}{supp.\thesection}

In this supplementary material, we provide additional details on our B-spline implementation (\cref{section:bpline_impl_details}), \textit{undistorted} microbolometer image model (\cref{section:undistorted_microbolo_model}), discrete approximation of motion blur (\cref{section:discrete_motion_blur}), mapping optimization (\cref{section:mapping_opt_details}), modified gradient accumulation (\cref{section:modified_gradient_accum}), and trajectory transformations applied after IMU initialization and IMU updates (\cref{section:imu_init_and_update_details}). We also provide additional details on the dataset we run on (\cref{section:dataset_details}) and the parameter values we use in all runs (\cref{section:full_parameter_values}). Moreover, we provide supplementary results in the form of an ablation study (\cref{section:ablation_details}) and further timing analysis (\cref{section:extended_timing}). Finally, we provide a broader discussion of the limitations of our method (\cref{section:limits}).

\section{B-Spline Implementation Details}
\label{section:bpline_impl_details}

We introduce a standalone package for B-spline optimization in PyTorch. The package supports B-splines in $\mathbb{R}^d$ and $\text{SO}(3)$. This section describes our trajectory implementation specifically, but the expressions involving position can generalize to $\mathbb{R}^d$ (and we exploit this to implement the FPN components with splines in $\mathbb{R}^{wh}$ and $\mathbb{R}$). The functionality described in \cref{section:bspline_ext_details} for extending the B-splines, fitting them to predictions, and making constant velocity predictions, is also included in the package (but IMU integration is implemented separately).

\subsection{B-spline Formulation}
\label{section:bpline_formulation}

As stated in \cref{section:bspline_traj_and_imu_loss}, we represent the continuous-time trajectory using two separate uniform B-splines for position $\mathbf{p}(t)\in\mathbb{R}^3$ and rotation $\mathbf{R}(t)\in\text{SO}(3)$. Specifically, we use a non-cumulative B-spline for the positions, and a cumulative B-spline for the rotations. At time $t\in[t_i, t_{i+1})$ , the position and rotation are given as:
\begin{equation}
\begin{gathered}
    \mathbf{p}(t) = \begin{bmatrix} \mathbf{p}_i & \mathbf{p}_{i+1} \!\!\! & \cdots & \!\!\! \mathbf{p}_{i+k-1} \end{bmatrix} \mathbf{M}^{(k)} \mathbf{u} = \mathbf{P}_i^{(k)} \mathbf{M}^{(k)} \mathbf{u}\\
    \mathbf{R}(t) = \mathbf{R}_i \prod_{j=1}^{k-1} \text{Exp} \left( \lambda_j(t) \, \text{Log} \left( \mathbf{R}_{i+j-1}^{-1} \mathbf{R}_{i+j} \right) \right) \\
    \boldsymbol{\lambda}(t) = \mathbf{M}^{*(k)}\mathbf{u}, \mathbf{u}=\begin{bmatrix} 1 & u \!\!\! & \cdots &  \!\!\! u^{k-1} \end{bmatrix}^{\text{T}}, u=\frac{t-t_i}{\Delta t_c} 
\end{gathered}
\label{equation:traj_spline}
\end{equation}
where $k$ is the order of the splines, $\{\mathbf{p}_i\}_{i=0}^{N_c-1}$ and $\{\mathbf{R}_i\}_{i=0}^{N_c-1}$ are the position and rotation control points, $\{t_i\}_{i=0}^{N_c-1}$ are the knots, $\Delta t_c=t_{i+1}-t_i \ \forall i$ is the knot interval, and $\mathbf{M}^{(k)}$ and $\mathbf{M}^{*(k)}$ are the non-cumulative and cumulative blending matrices, respectively. Note that for uniform B-splines (those with constant $\Delta t_c$) both $\mathbf{M}^{(k)}$ and $\mathbf{M}^{*(k)}$ are constant and determined solely by the spline order $k$ \cite{sommer_efficient_2020}.

\subsection{B-spline Derivatives}
\label{section:bpline_derivatives}

Given the definitions of the position and rotation splines in \cref{equation:traj_spline}, the linear velocity $\mathbf{v}(t)$ and acceleration $\mathbf{a}(t)$ can be computed as:
\begin{equation}\label{equation:linear_vel_accel}
\begin{gathered}
\mathbf{v}(t) = \mathbf{P}_i^{(k)} \mathbf{M}^{(k)} \dot{\mathbf{u}}, \ \ \mathbf{a}(t) = \mathbf{P}_i^{(k)} \mathbf{M}^{(k)} \ddot{\mathbf{u}}  \\
\dot{\mathbf{u}}=\frac{1}{\Delta t_c} \begin{bmatrix} 0 & 1 & 2u \!\!\! & \cdots & \!\!\! (k-1)u^{k-2} \end{bmatrix}^{\text{T}} \\
\ddot{\mathbf{u}}=\frac{1}{\Delta t_c^2} \begin{bmatrix} 0 & 0 & 2 & 6u \!\!\! & \cdots & \!\!\! (k-1)(k-2)u^{k-3} \end{bmatrix}^{\text{T}}
\end{gathered}
\end{equation}
and we can compute the angular velocity $\boldsymbol{\omega}(t)$ directly by applying the following efficient recursive expression from $j=1$ to $j=k$ \cite{sommer_efficient_2020}:
\begin{equation}
\begin{gathered}
\boldsymbol{\omega}^{(j)}(t) = \left(\mathbf{A}_{j-1}^i(t)\right)^{-1} \boldsymbol{\omega}^{(j-1)}(t) + \dot{\lambda}_{j-1}(t) \ \mathbf{d}_{j-1}^i \\
\mathbf{A}_{j}^i(t) = \text{Exp}\left(\lambda_j(t)\mathbf{d}_j^i\right), \ \mathbf{d}_j^i=\text{Log} \left( \mathbf{R}_{i+j-1}^{-1} \mathbf{R}_{i+j} \right) \\
\dot{\boldsymbol{\lambda}}(t)=\mathbf{M}^{*(k)}\dot{\mathbf{u}}, \ \boldsymbol{\omega}^{(1)}(t)=\mathbf{0}
\end{gathered}
\label{equation:angular_vel}
\end{equation}

\subsection{On-Manifold Parameter Updates, Efficient Forward Passes, \& Analytical Backward Passes for SO(3) B-Splines}
\label{section:on_manifold}

One common approach to optimizing rotations in PyTorch is to store them in the minimal axis-angle format, apply the exponential and logarithm maps to convert to and from rotation matrices as needed, and rely on PyTorch's default tools for automatic differentiation and optimization. Storing the parameters as axis-angle vectors ensures they remain valid rotations, but the standard PyTorch backward passes through the exponential and logarithm maps include numerically unstable terms \cite{teed2021tangent}. It is more stable and accurate to compute tangent space gradients through custom backward passes and perform on-manifold parameter updates \cite{teed2021tangent}. PyPose \cite{wang2023pypose} and LieTorch \cite{teed2021tangent} are PyTorch libraries that provide this functionality.

In our package, we implement the rotation spline to return a \texttt{pypose.SO3} \cite{wang2023pypose} or \texttt{lietorch.SO3} \cite{teed2021tangent} object when it is evaluated (\cref{equation:traj_spline}). In either case, this ensures that the gradients for the returned rotations are computed in the tangent space. Our custom backward passes, described below, also compute gradients in the tangent space, and we perform on-manifold updates to the rotation control points in optimization. That is, an update step for control point $\mathbf{R}_i$ is performed as $\mathbf{R}_i\leftarrow\text{Exp}(\boldsymbol{\delta}_i)\mathbf{R}_i$ where $\boldsymbol{\delta}_i$ is the optimizer increment computed from the tangent space gradient of $\mathbf{R}_i$.

We implement three different versions of the forward passes for evaluating the rotation spline \cref{equation:traj_spline} and its angular velocity \cref{equation:angular_vel}: one using PyPose \cite{wang2023pypose}, one using LieTorch \cite{teed2021tangent}, and one using multi-threaded C++ functions that run on the CPU and are called through Python bindings. We leave the backward passes for the LieTorch and PyPose versions to be handled automatically by their respective libraries. For the C++ version, we implement custom backward passes utilizing the efficient algorithms given in \cite{sommer_efficient_2020} for computing the Jacobians of the rotations and angular velocities with respect to the control points. To avoid redundant calculations, these Jacobians are computed in the forward pass and stored for the backward pass if gradients are needed.

\begin{table}[]
\centering
\caption{Timing comparison between different versions of our rotation spline.}
\label{tab:spline_timing}
\resizebox{\columnwidth}{!}{%
\begin{tabular}{@{}lcccccc@{}}
\toprule
\multirow{2}{*}{\textbf{Version}} & \multirow{2}{*}{\textbf{Device}} & \multirow{2}{*}{\textbf{\begin{tabular}[c]{@{}c@{}}Number of\\ Timestamps\\ Evaluated\end{tabular}}} & \multicolumn{2}{c}{\textbf{\begin{tabular}[c]{@{}c@{}}Rotation Loss\\ Compute Time (ms)\end{tabular}}} & \multicolumn{2}{c}{\textbf{\begin{tabular}[c]{@{}c@{}}Ang. Vel. Loss\\ Compute Time (ms)\end{tabular}}} \\ \cmidrule(l){4-7} 
                                  &                                  &                                                                                                      & \textbf{Forward}                                    & \textbf{Backward}                                    & \textbf{Forward}                                     & \textbf{Backward}                                    \\ \midrule
PyPose                            & CPU                              & 1k                                                                                                   & 5.45                                                & 4.97                                                 & 5.64                                                 & 5.71                                                 \\
LieTorch                          & CPU                              & 1k                                                                                                   & 1.68                                                & 1.39                                                 & 1.82                                                 & 1.48                                                 \\
C++                               & CPU                              & 1k                                                                                                   & \textbf{0.59}                                       & 0.41                                                 & \textbf{0.51}                                        & \textbf{0.27}                                        \\
PyPose                            & GPU                              & 1k                                                                                                   & 14.30                                               & 13.74                                                & 15.44                                                & 15.55                                                \\
LieTorch                          & GPU                              & 1k                                                                                                   & 3.11                                                & 2.12                                                 & 4.28                                                 & 2.28                                                 \\
C++                               & GPU                              & 1k                                                                                                   & 0.81                                                & \textbf{0.38}                                        & 0.73                                                 & 0.30                                                 \\ \midrule
PyPose                            & CPU                              & 100k                                                                                                 & 34.70                                               & 89.19                                                & 30.76                                                & 91.55                                                \\
LieTorch                          & CPU                              & 100k                                                                                                 & 44.99                                               & 80.67                                                & 43.78                                                & 85.54                                                \\
C++                               & CPU                              & 100k                                                                                                 & 16.65                                               & 14.47                                                & 11.16                                                & 3.20                                                 \\
PyPose                            & GPU                              & 100k                                                                                                 & 14.87                                               & 15.63                                                & 15.25                                                & 17.43                                                \\
LieTorch                          & GPU                              & 100k                                                                                                 & \textbf{3.93}                                       & 2.37                                                 & \textbf{4.20}                                        & 2.38                                                 \\
C++                               & GPU                              & 100k                                                                                                 & 18.90                                               & \textbf{1.22}                                        & 33.89                                                & \textbf{1.11}                                        \\ \bottomrule
\end{tabular}%
}
\end{table}

\Cref{tab:spline_timing} shows a timing comparison between the three versions we implemented, operating with a NVIDIA RTX A6000 GPU and an AMD Ryzen 9 5950X CPU. The comparison shows the time taken for the forward and backward pass of computing a rotation loss (similar to \cref{equation:fit_problems}) and an angular velocity loss (similar to \cref{equation:imu_loss}, without the accelerometer term). The table shows results for the splines being evaluated at 1k timestamps and 100k timestamps, with the control points stored on the CPU or GPU. Note that when our C++ implementation is used with the control points stored on the GPU, data is transferred to the CPU for the forward pass only. The results indicate that the C++ version, with control points stored on the CPU, is the most efficient when the spline is evaluated at a relatively small number of timestamps, while the LieTorch version, with control points stored on the GPU, is the fastest with a large number of timestamps. Therefore, we use the C++ version in SLAM, with the trajectory control points and the IMU bias parameters stored on the CPU, and the LieTorch version in offline refinement, with all parameters stored on the GPU.

Additionally, as we generally observed LieTorch operations and backward passes to be more efficient than PyPose, we configure the rotation spline to return a \texttt{lietorch.SO3} object when evaluated in SLAM and offline refinement (regardless of whether the actual spline evaluation was performed with LieTorch or our C++ implementation). 

\subsection{Sparse Gradients and Optimization}
\label{section:sparse_opt}

For each spline, we store the entire set of control points as an individual PyTorch \texttt{Parameter}. To ensure that only active control points are updated at each iteration, our spline implementations return sparse gradients and we use PyTorch's \texttt{SparseAdam} optimizer to update them. If we were to return dense gradients and use the standard \texttt{Adam} optimizer, the inactive control point gradients would be zero, the corresponding moments in the optimizer state would be falsely updated using these zero-valued gradients, and the inactive control points themselves would be falsely updated (if their moments are non-zero from past active iterations), despite taking no part in the current loss calculation. \texttt{SparseAdam}, by contrast, only updates the moments and parameters indexed by the sparse gradient it receives. In our implementation, this is essentially equivalent to, though much more efficient than, storing the control points as individual PyTorch \texttt{Parameter}s and updating them with separate \texttt{Adam} optimizers\footnote{The current PyTorch implementation of \texttt{SparseAdam} uses a single step counter to compute the bias correction. This quickly reduces to \texttt{Adam} \textit{without} bias correction for new control points. We tested a custom implementation with an element-wise step counter but found empirically that this harmed performance.}. 

\subsection{B-Spline Extension Details}
\label{section:bspline_ext_details}

\paragraph{Duplication \& Fitting to Predictions}
As explained in \cref{section:bspline_ext}, the position and rotation splines can only be evaluated for $t \in [t_0, t_0 + (N_c - k + 1) \Delta t_c)$ and control points must be added when a new image arrives. The number of new control points needed to accommodate image $l$ is:
\begin{equation}
    N_{\text{new}} = \lfloor \frac{t_{\text{max}, l} - t_0}{\Delta t_c} \rfloor + k - N_c
\label{equation:num_new_control_points}
\end{equation}
These new control points are initialized by duplicating $\mathbf{p}_{N_c-1}$ and $\mathbf{R_{N_c-1}}$.

After a set number of keyframes, we additionally compute $N_q$ predicted poses at times $\{t_{\text{pred}, q}\}_{q=0}^{N_q-1}$ evenly spaced over $[t'_{0, 0, l-1}, t_{\text{max}, l}]$ and fit the active sets of position and rotation control points,
\begin{equation}
\begin{aligned}
\mathcal{P}&=\{\mathbf{p}_i\}_{i= \lfloor (t'_{0, 0, l-1} - t_0) / \Delta t_c \rfloor}^{N_c + N_{\text{new}}-1} \\ 
\mathcal{R}&=\{\mathbf{R}_i\}_{i= \lfloor (t'_{0, 0, l-1} - t_0) / \Delta t_c \rfloor}^{N_c + N_{\text{new}}-1}
\end{aligned}
\end{equation}
to these predictions. Specifically, we separately solve the following optimization problems:
\begin{equation}
\begin{aligned}
    &\underset{\mathcal{P}}{\text{arg min}} \sum_{q=0}^{N_q-1} \left\| \mathbf{p}(t_{\text{pred}, q})- \mathbf{p}_{\text{pred}, q}\right\|_2^2 \\
    &\underset{\mathcal{R}}{\text{arg min}} \sum_{q=0}^{N_q-1} \left\| \text{Log}\left(\left(\mathbf{R}(t_{\text{pred}, q})\right)^{-1}\mathbf{R}_{\text{pred}, q} \right) \right\|_2^2
\end{aligned}
\label{equation:fit_problems}
\end{equation}
where $\{ \mathbf{p}_{\text{pred}, q} \}_{q=0}^{N_q-1}$ and $\{ \mathbf{R}_{\text{pred}, q} \}_{q=0}^{N_q-1}$ are the predicted positions and rotations. $N_q$ is set larger than the number of active control points ($N_q > N_c + N_{\text{new}}- \lfloor (t'_{0, 0, l-1} - t_0) / \Delta t_c \rfloor$) to ensure the problems are overdetermined. We use the PyPose \cite{wang2023pypose} implementation of Levenberg–Marquard to solve these problems within the PyTorch framework.

Note that it is necessary to fit the control points to the predictions because B-spline control points do not generally sit on the spline itself and therefore the value of the spline cannot be trivially set directly.

\paragraph{Constant Velocity Predictions}
Prior to IMU initialization, the predictions are made assuming constant velocity. The instantaneous velocities of the splines (Eqs. \ref{equation:linear_vel_accel} and \ref{equation:angular_vel}) can be noisy, especially prior to IMU initialization. Therefore, we compute finite difference estimates of the linear and angular velocities as follows:
\begin{equation}
\begin{split}
    \bar{\mathbf{v}}_{\text{C}, \text{avg}}^{\text{W}}&(t'_{0, 0, l-1}) = \\
    & \frac{\bar{\mathbf{p}}_{\text{C}}^{\text{W}}(t'_{0, 0, l-1}) - \bar{\mathbf{p}}_{\text{C}}^{\text{W}}(t'_{0, 0, l-1}-T_I)}{T_I} \\
    \boldsymbol{\omega}_{\text{C}, \text{avg}}^{\text{C}}&(t'_{0, 0, l-1}) = \\
    & \frac{\text{Log}\left( \left(\mathbf{R}_{\text{C}}^{\text{W}}(t'_{0, 0, l-1}-T_I)\right)^{-1}\mathbf{R}_{\text{C}}^{\text{W}}(t'_{0, 0, l-1}) \right)}{T_I}
\end{split}
\end{equation}
where $T_I$ is the average frame period. The predicted position and rotation at time $t_{\text{pred}, q}$ are then computed as:
\begin{equation}
\begin{split}
    \bar{\mathbf{p}}&_{\text{C}, \text{pred}, q}^{\text{W}} = \\
    & \bar{\mathbf{p}}_{\text{C}}^{\text{W}}(t'_{0, 0, l-1}) + (t_{\text{pred}, q} - t'_{0, 0, l-1})  \ \bar{\mathbf{v}}_{\text{C}, \text{avg}}^{\text{W}}(t'_{0, 0, l-1}) \\
    \mathbf{R}&_{\text{C}, \text{pred}, q}^{\text{W}} = \\
    & \mathbf{R}_{\text{C}}^{\text{W}}(t'_{0, 0, l-1}) \text{Exp}\left((t_{\text{pred}, q} - t'_{0, 0, l-1}) \ \boldsymbol{\omega}_{\text{C}, \text{avg}}^{\text{C}}(t'_{0, 0, l-1})\right)
\end{split}
\end{equation}

\paragraph{IMU Integration Based Predictions}
After IMU initialization, the predictions are made through IMU integration. Let $b$ denote the index of an IMU measurement taken exactly at $t'_{0, 0, l-1}$ and $f$ denote the index of the last IMU measurement taken before the prediction time $t_{\text{pred}, q}$. Note that $\tilde{\boldsymbol{\omega}}_{\text{I}, b}^{\text{I}}$ and $\tilde{\mathbf{a}}_{\text{I}, b}^{\text{I}}$ are interpolated using the IMU measurements taken immediately before and after $t'_{0, 0, l-1}$. We compute the position and rotation at the time of the final IMU measurement $t_{\text{IMU}, f}$ by integrating the IMU measurements from $t_{\text{IMU}, b}$ to $t_{\text{IMU}, f}$ as follows:
\begin{equation}
\begin{split}
    \mathbf{R}_{\text{I}, \text{IMU}, f}^{\text{W}} & = \mathbf{R}_{\text{I}}^{\text{W}}(t'_{0, 0, l-1}) \prod_{m=b}^{f-1} \text{Exp} \left( (\tilde{\boldsymbol{\omega}}_{\text{I}, m}^{\text{I}} - \mathbf{b_g}) \Delta t_m \right) \\
    \mathbf{v}_{\text{I}, \text{IMU}, f}^{\text{W}} & = \mathbf{v}_{\text{I}}^{\text{W}}(t'_{0, 0, l-1}) + \sum_{m=b}^{f-1} G \hat{\mathbf{g}}^{\text{W}} \Delta t_m \\
    & + \sum_{m=b}^{f-1} \mathbf{R}_{\text{I}, \text{IMU}, m}^{\text{W}} (\tilde{\mathbf{a}}_{\text{I}, m}^{\text{I}} - \mathbf{b}_a) \Delta t_m \\
    \mathbf{p}_{\text{I}, \text{IMU}, f}^{\text{W}} & = \mathbf{p}_{\text{I}}^{\text{W}}(t'_{0, 0, l-1}) + \sum_{m=b}^{f-1} \mathbf{v}_{\text{I}, \text{IMU}, m}^{\text{W}} \Delta t_m \\
    & + \sum_{m=b}^{f-1} \frac{1}{2}G \hat{\mathbf{g}}^{\text{W}} \Delta t_m^2 \\
    & + \sum_{m=b}^{f-1} \frac{1}{2}\mathbf{R}_{\text{I}, \text{IMU}, m}^{\text{W}} (\tilde{\mathbf{a}}_{\text{I}, m}^{\text{I}} - \mathbf{b}_a) \Delta t^2_m
\end{split}
\end{equation}
where $\Delta t_m = t_{\text{IMU}, m+1} - t_{\text{IMU}, m}$. Note that in this case we do use the instantaneous linear velocity $\mathbf{v}_{\text{I}}^{\text{W}}(t'_{0, 0, l-1})$  (\cref{equation:linear_vel_accel}), rather than a finite difference estimate. The predicted position and rotation at time $t_{\text{pred}, q}$ are then computed as:
\begin{equation}
\begin{split}
    \mathbf{p^{\text{W}}_{\text{I}, \text{pred}, q}} &= \mathbf{p}_{\text{I}, \text{IMU}, f}^{\text{W}} + \mathbf{v}_{\text{I}, \text{IMU}, f}^{\text{W}}\Delta t_f + \frac{1}{2}G \hat{\mathbf{g}}^{\text{W}} \Delta t^2_f \\
    & + \frac{1}{2}\mathbf{R}_{\text{I}, \text{IMU}, f}^{\text{W}} (\tilde{\mathbf{a}}_{\text{I}, f}^{\text{I}} - \mathbf{b}_a) \Delta t^2_f\\
    \mathbf{R}^{\text{W}}_{\text{I}, \text{pred}, q} & = \mathbf{R}_{\text{I}, \text{IMU}, f}^{\text{W}} \text{Exp} \left( \tilde{\boldsymbol{\omega}}_{\text{I}, f}^{\text{I}}\Delta t_f \right) \\
\end{split}
\end{equation}
where $\Delta t_f = t_{\text{pred}, q} - t_{\text{IMU}, f}$.

\paragraph{Optimizer State Management}

When the B-splines are extended, the control point \texttt{Parameter}s are re-initialized as the existing control points concatenated with the new ones. The optimizer states corresponding to the new control points are set to zero while the states corresponding to the existing control points are retained. Note that in the fitting process (\cref{equation:fit_problems}) the active sets of control points, $\mathcal{P}$ and $\mathcal{R}$, include existing control points that have already been optimized through tracking and mapping. We initially restricted the B-spline extension process to only fit the $N_{\text{new}}$ control points to the pose predictions, to avoid manipulating control points that may have already been well optimized. However, this seemingly overfit the new control points, leading to a poor initialization for tracking. It is also unclear whether the optimizer states corresponding to these modified control points should be zeroed. In practice, we have found it best to leave these optimizer states unchanged. 

\section{\textit{Undistorted} Microbolometer Image Model}
\label{section:undistorted_microbolo_model}

The undistortion of $\tilde{\mathbf{N}}'_l$ can be written as follows:
\begin{equation}
\begin{split}
[\tilde{\mathbf{N}}_l]_{y_p, x_p} &  = \text{undistort}(\tilde{\mathbf{N}}'_l, x_p, y_p) \\
& = f_{\text{interp}}(\tilde{\mathbf{N}}'_l, \text{map}_x(x_p, y_p), \text{map}_y(x_p, y_p)) 
\end{split}
\label{equation:image_undistort}
\end{equation}
where $\text{map}_x(x_p, y_p)$ and $\text{map}_y(x_p, y_p)$ map pixel coordinates $(x_p, y_p)$ in the undistorted image $\tilde{\mathbf{N}}_l$ to fractional pixel coordinates in $\tilde{\mathbf{N}}'_l$, and $f_{\text{interp}}$ denotes bilinear interpolation. $\text{map}_x(x_p, y_p)$ and $\text{map}_y(x_p, y_p)$ are precomputed from the original camera intrinsic matrix, distortion parameters, and the new camera matrix. Bilinear interpolation is used here to estimate the value of $\tilde{\mathbf{N}}'_l$ at the fractional pixel coordinates and it can written as:
\begin{equation}
f_{\text{interp}}(\mathbf{Q}, x, y) = \sum_{i\in\{0, 1\}}\sum_{j\in\{0, 1\}} w_{ij}[\mathbf{Q}]_{y_f+j, x_f+i}
\label{equation:bilinear_interp}
\end{equation}
where $x_f=\lfloor x \rfloor$, $y_f=\lfloor y \rfloor$, and $\{w_{00}, w_{01}, w_{10},w_{11}\}$ are weights computed based the relative proximity of $(x, y)$ to the four neighboring integer coordinates. Note that these weights sum to 1.

Let $x=\text{map}_x(x_p, y_p)$ and $y=\text{map}_y(x_p, y_p)$. We can then write:
\begin{equation}
\begin{split}
[\tilde{\mathbf{N}}_l]_{y_p, x_p} &= \\
& \sum_{i\in\{0, 1\}}\sum_{j\in\{0, 1\}} w_{ij} \ \tilde{n}'_{x_f+i, y_f+j}(t'_{x_f+i, y_f+j, l})
\end{split}
\label{equation:undistort_as_weighted_sum}
\end{equation}
Given that the four neighboring pixels span only two rows, their readout times differ only slightly. Therefore, we define the effective readout time of the pixel $(x_p, y_p)$ in the undistorted image as:
\begin{equation}
t_{x_p, y_p, l}=\sum_{i\in\{0, 1\}}\sum_{j\in\{0, 1\}} w_{ij} \ t'_{x_f+i, y_f+j, l}
\label{equation:effective_readout}
\end{equation}
With this effective readout time, we can then write $\tilde{[\mathbf{N}}_l]_{y_p, x_p} = \tilde{n}_{x_p, y_p}(t_{x_p, y_p, l})$ and approximate \cref{equation:undistort_as_weighted_sum} as:
\begin{equation}
[\tilde{\mathbf{N}}_l]_{y_p, x_p} \approx \sum_{i\in\{0, 1\}}\sum_{j\in\{0, 1\}} w_{ij} \ \tilde{n}'_{x_f+i, y_f+j}(t_{x_p, y_p, l})
\end{equation}
Next, by expanding $\ \tilde{n}'_{x_f+i, y_f+j}(t_{x_p, y_p, l})$ and distributing the sum, we can write $\tilde{n}_{x_p, y_p}(t_{x_p, y_p, l})$ in the form of Eqs. \ref{equation:final_pixel} and \ref{equation:just_blur}:
\begin{equation}
    \tilde{n}_{x_p, y_p}(t_{x_p, y_p, l}) = \tilde{m}_{x_p, y_p}(t_{x_p, y_p, l}) + \tilde{o}_{x_p, y_p}(t_{x_p, y_p, l})
\end{equation}
\begin{equation}
\begin{split}
\tilde{m}_{x_p, y_p}(t&_{x_p, y_p, l}) = \\ 
&\frac{1}{\tau}\int_{-\infty}^{t_{x_p, y_p, l}} \exp\left({\frac{s-t_{x_p, y_p, l}}{\tau}}\right) \tilde{p}_{x_p, y_p}(s)ds \label{equation:just_blur_undistorted}
\end{split}
\end{equation}
where $\tilde{m}_{x_p, y_p}(t)$, $\tilde{o}_{x_p, y_p}(t)$ and $\tilde{p}_{x_p, y_p}(t)$ are defined as:
\begin{align}
    \tilde{m}_{x_p, y_p}(t) = \sum_{i\in\{0, 1\}}\sum_{j\in\{0, 1\}} w_{ij} \ \tilde{m}'_{x_f+i, y_f+j}(t) \\
    \tilde{o}_{x_p, y_p}(t) = \sum_{i\in\{0, 1\}}\sum_{j\in\{0, 1\}} w_{ij} \ \tilde{o}'_{x_f+i, y_f+j}(t) \\
    \tilde{p}_{x_p, y_p}(t) = \sum_{i\in\{0, 1\}}\sum_{j\in\{0, 1\}} w_{ij} \ \tilde{p}'_{x_f+i, y_f+j}(t)
\end{align}

In summary, the undistorted image can be written as $[\tilde{\mathbf{N}}_l]_{y_p, x_p} = \tilde{n}_{x_p, y_p}(t_{x_p, y_p, l})$ where the individual pixel values $\tilde{n}_{x_p, y_p}(t_{x_p, y_p, l})$ can be written in the same form as Eqs. \ref{equation:final_pixel} and \ref{equation:just_blur}, but the readout time is determined by \cref{equation:effective_readout}.

Note that the effective readout time in \cref{equation:effective_readout} can be easily computed using the same undistortion operation applied to the images. Let $\mathbf{T}'_\text{roll}$ be a lookup table of the readout times in the distorted image, relative to $t'_{0, 0, l}$. That is, $[\mathbf{T}'_\text{roll}]_{y_p, x_p} = x_p\Delta t_{\text{pix}} + y_p w \Delta t_{\text{pix}}$. If we expand \cref{equation:effective_readout}:
\begin{equation}
\begin{split}
t_{x_p, y_p, l} &=\sum_{i\in\{0, 1\}}\sum_{j\in\{0, 1\}} w_{ij} \ t'_{0, 0, l} \\
& \quad + \sum_{i\in\{0, 1\}}\sum_{j\in\{0, 1\}} w_{ij} \ [\mathbf{T}'_\text{roll}]_{y_{f+j}, {x_f+i}} \\
& = t'_{0, 0, l} + \sum_{i\in\{0, 1\}}\sum_{j\in\{0, 1\}} w_{ij} \ [\mathbf{T}'_\text{roll}]_{y_{f+j}, {x_f+i}}
\end{split}
\end{equation}
it becomes clear from comparison to Eqs. \ref{equation:image_undistort} and \ref{equation:bilinear_interp} that this can written as:
\begin{equation}
\begin{split}
t_{x_p, y_p, l} & = t'_{0, 0, l} + \text{undistort}(\mathbf{T}'_\text{roll}, x_p, y_p) \\
& = t'_{0, 0, l} +  [\mathbf{T}_\text{roll}]_{y_p, x_p}
\end{split}
\end{equation}

Finally, note that we also downsample the images prior to undistortion. We downsample by block averaging, using a block size the image dimensions are divisible by. Following a process similar to the preceding discussion on undistortion, it can be shown that this block averaging approximately preserves the image formation model but with new effective readout times. In this case, the new readout times follow the original expression given in \cref{equation:rs_readout}, but with $\Delta t_{\text{pix}}$ multiplied by the square of the block size.

\section{Discrete Approximation of Motion Blur}
\label{section:discrete_motion_blur}

As explained in \cref{section:deg_aware_rendering}, to compute $[\mathbf{M}_l]_{y_p, x_p} = m_{x_p, y_p}(t_{x_p, y_p, l})$ we rasterize $N_r$ images at times $\{t_r\}_{r=0}^{N_r-1}$ evenly spaced over $[t_{\text{max}, l}-T_b, t_{\text{max}, l}]$ where $t_{\text{max}, l}$ is defined as:
\begin{equation}
    t_{\text{max}, l} = t'_{0, 0, l} +  \max(\mathbf{T}_\text{roll})
\end{equation}

Focusing on a single pixel, this means we have $N_r$ samples $\{p_{x_p, y_p}(t_r)\}_{r=0}^{N_r-1}$ and need to compute an estimate of the blurry pixel value $m_{x_p, y_p}(t_{x_p, y_p, l})$ with a discrete approximation of \cref{equation:just_blur_undistorted}. In the following derivation we assume the most general case: the effective readout time falls between two samples, i.e., $t_n < t_{x_p, y_p, l} < t_{n+1}$ for some $n$ where $0 \leq n < N_r-1$. 

To compute $m_{x_p, y_p}(t_{x_p, y_p, l})$ we treat $p_{x_p, y_p}(t_r)$ as linear between samples, and break \cref{equation:just_blur_undistorted} up into a sum of integrals over consecutive samples. Let $t_a$ and $t_b$ be consecutive sample times with $t_b \leq t_n$.  Dropping the pixel coordinate subscripts for readability, we can write the portion of the integral in \cref{equation:just_blur_undistorted} from $t_a$ to $t_b$ as:
\begin{equation}
\begin{split}
    &\frac{1}{\tau}\int_{t_a}^{t_b} \exp\left(\frac{s-t_l}{\tau}\right) p(s) ds = \\
    &\frac{1}{\tau}\int_{t_a}^{t_b} \exp\left(\frac{s-t_l}{\tau}\right) \left( p(t_a) + v_{a, b} \ (s-t_a) \right) ds
\label{equation:integral_segment}
\end{split}
\end{equation}
where $v_{a, b}$ is the slope of $p(t)$ between $t_a$ and $t_b$:
\begin{equation}
\begin{aligned}
    v_{a, b} &= \frac{p(t_b) - p(t_a)}{t_b - t_a}=\frac{p(t_b) - p(t_a)}{\Delta t_r}, \\
    \Delta t_r &= \frac{T_b}{N_r - 1}
\end{aligned}
\label{equation:segment_slope}
\end{equation}
The result of \cref{equation:integral_segment} is:
\begin{equation}
\begin{split}
& \left[ \exp\left( \frac{s-t_l}{\tau}\right) \left(p(t_a) - v_{a, b} \ (t_a-s+\tau)\right) \right]_{t_a}^{t_b} = \\
& \left[ \exp\left( \frac{s-t_l}{\tau}\right) \left((p(t_a) + v_{a, b} \ (s-t_a)) - v_{a, b} \ \tau\right) \right]_{t_a}^{t_b} = \\
& \left[ \exp\left( \frac{s-t_l}{\tau}\right) \left(p(s) - v_{a, b} \ \tau\right) \right]_{t_a}^{t_b} 
\end{split}
\label{equation:integral_result}
\end{equation}
Evaluating \cref{equation:integral_result}, substituting in \cref{equation:segment_slope}, letting $E_a=\exp \left(\frac{t_a-t_l}{\tau}\right)$ and $E_b=\exp \left(\frac{t_b-t_l}{\tau}\right)$, and grouping terms by $p(t_a)$ and $p(t_b)$ we get:
\begin{equation}
\begin{split}
    &\frac{1}{\tau}\int_{t_a}^{t_b} \exp\left(\frac{s-t_l}{\tau}\right) p(s) ds = \\
    & \quad\quad \ p(t_a)(E_b \frac{\tau}{\Delta t_r}-E_a(1+\frac{\tau}{\Delta t_r})) \\
    & \quad\quad + \ p(t_b)(E_b(1-\frac{\tau}{\Delta t_r})+E_a \frac{\tau}{\Delta t_r})
\label{equation:grouped_full_segment}
\end{split}
\end{equation}

Next, because $t_n < t_l < t_{n+1}$, the portion of the integral in \cref{equation:just_blur_undistorted} from $t_n$ to $t_l$ is computed as:
\begin{equation}
\begin{split}
&\frac{1}{\tau}\int_{t_n}^{t_l} \exp\left(\frac{s-t_l}{\tau}\right) p(s) ds = \\
& \left[ \exp\left( \frac{s-t_l}{\tau}\right) \left(p(s) - v_{n, n+1} \ \tau\right) \right]_{t_n}^{t_l} =  \\
& \ p(t_n)\left(1-\frac{t_l-t_n}{\Delta t_r}+\frac{\tau}{\Delta t_r}-E_n (1+\frac{\tau}{\Delta t_r})\right) \ \\
    & + p(t_{n+1})\left(\frac{t_l-t_n}{\Delta t_r}-\frac{\tau}{\Delta t_r}+E_n \frac{\tau}{\Delta t_r}\right)
\label{equation:grouped_partial_segment}
\end{split}
\end{equation}

Putting this together we can write:
\begin{equation}
    \frac{1}{\tau}\int_{t_{\text{max}, l}-T_b}^{t_l} \exp\left(\frac{s-t_l}{\tau}\right) p(s) ds = \sum_{r=0}^{N_r-1} (h_r + d_r) p(t_r)
\label{equation:combined_segments}
\end{equation}
where $h_r$ and $d_r$ are set based on the grouped terms in Eqs. \ref{equation:grouped_full_segment} and \ref{equation:grouped_partial_segment}, respectively. Specifically, using $j=\lfloor \frac{t_l-t_r}{\Delta t_r} \rfloor$, $h_r$ and $d_r$ are defined as:
\begin{equation}
    h_r = 
\begin{cases}
0 & \text{if } j < 0 \\
0 & \text{if } j=0, r=0 \\
E_r(1-\frac{\tau}{\Delta t_r})+E_{r-1} \frac{\tau}{\Delta t_r} & \text{if } j=0, r\neq0 \\
E_{r+1} \frac{\tau}{\Delta t_r}-E_r(1+\frac{\tau}{\Delta t_r}) & \text{if } j>0,r=0 \\
\frac{\tau}{\Delta t_r} (E_{r-1} - 2 E_r + E_{r+1}) & \text{if } j > 0,r\neq0
\end{cases}
\end{equation}

\begin{equation}
    d_r = 
\begin{cases}
\frac{t_l-t_{r-1}}{\Delta t_r}-\frac{\tau}{\Delta t_r}+E_{r-1} \frac{\tau}{\Delta t_r} & \text{if } j = -1 \\
1-\frac{t_l-t_r}{\Delta t_r}+\frac{\tau}{\Delta t_r}-E_r (1+\frac{\tau}{\Delta t_r}) & \text{if } j=0 \\
0 & \text{else}
\end{cases}
\end{equation}
Note that the cases involving $j$ are used to determine whether the sample $p(t_r)$ contributes to a full segment (\cref{equation:grouped_full_segment}), partial segment (\cref{equation:grouped_partial_segment}), or both. For $r\neq0$:
\begin{itemize}
    \item $j = -1$ implies $r$ is the right endpoint of the partial segment ($r=n+1$ in \cref{equation:grouped_partial_segment}), only
    \item $j=0$ implies $r$ is the left endpoint of the partial segment ($r=n$ in  \cref{equation:grouped_partial_segment}) and the right endpoint of the last full segment ($r=b$ in \cref{equation:grouped_full_segment})
    \item $j>0$ implies $r$ is an interior point, acting as both the left endpoint and right endpoint of two consecutive full segments ($r=a$ and $r=b$ in \cref{equation:grouped_full_segment})
\end{itemize}
In the special case of the first sample, $r=0$:
\begin{itemize}
    \item $j=0$ implies $r$ is the left endpoint of the partial segment ($r=n$ in  \cref{equation:grouped_partial_segment}), only
    \item $j>0$ implies $r$ is the left endpoint of a full segment ($r=a$ in \cref{equation:grouped_full_segment}), only
\end{itemize}

Given that the integral in \cref{equation:just_blur_undistorted} is unbounded, the result of the truncated integral in \cref{equation:combined_segments} may be significantly smaller than the true value. To approximately compensate for this we consider the case of constant $p(t)=p$. In this case, the result of the unbounded integral in \cref{equation:just_blur_undistorted} is $p$ and the result of \cref{equation:combined_segments} is:
\begin{equation}
\begin{split}
    \frac{p}{\tau}\int_{t_{\text{max}, l}-T_b}^{t_l} & \exp\left(\frac{s-t_l}{\tau}\right) ds = \\
    & p \left( 1 - \exp \left( \frac{t_{\text{max}, l}-T_b-t_l}{\tau} \right) \right)
\end{split}
\end{equation}
Therefore, we compute the pixel-wise correction factor as:
\begin{equation}
    g = \left( 1 - \exp \left( \frac{t_{\text{max}, l}-T_b-t_l}{\tau} \right) \right)^{-1}
\end{equation}

The final expression for $m_{x_p, y_p}(t_{x_p, y_p, l})$ is:
\begin{equation}
\begin{split}
& m_{x_p, y_p}(t_{x_p, y_p, l}) = \\
& \quad g_{x_p, y_p} \sum_{r=0}^{N_r-1} (h_{x_p, y_p, r} + d_{x_p, y_p, r}) p_{x_p, y_p}(t_r)
\end{split}
\end{equation}
where the pixel coordinate subscripts have been reintroduced for clarity.

Finally, we can construct the weight matrix $\mathbf{W}_r$ as:
\begin{equation}
    [\mathbf{W}_r]_{y_p, x_p} = g_{x_p, y_p} (h_{x_p, y_p, r} + d_{x_p, y_p, r})
\end{equation}
Pivotally, $[\mathbf{W}_r]_{y_p, x_p}$ is not dependent on the absolute readout time of the pixel $t_{x_p, y_p, l}$. Rather, $[\mathbf{W}_r]_{y_p, x_p}$ is only dependent on the relative time offsets between $t_{x_p, y_p, l}$ and the raster times $\{t_r\}_{r=0}^{N_r-1}$. Therefore, the weight matrices $\{\mathbf{W}_r\}_{r=0}^{N_r-1}$ can be precomputed.

\section{Mapping Optimization Details}
\label{section:mapping_opt_details}

In \cref{section:mapping} it was mentioned that $\mathcal{W}_I$ and $\mathcal{S}_I$ are disjoint sets of IMU measurements associated with $\mathcal{W}_K$ and $\mathcal{S}_K$. Specifically, the full set of IMU measurements used in a given mapping iteration $\mathcal{I}_{\text{map}}$ is the union of all sets of IMU measurements taken over $[\min(t_{0, 0, p}, t_{\text{max},l}-T_b), t_{\text{max}, l}]$ for each keyframe image $l$, where $p$ denotes the image index of the keyframe prior to $l$. The measurements are split into $\mathcal{W}_I$ or $\mathcal{S}_I$ depending on whether they came after or before $t_{\text{max},l}$ of the latest randomly sampled keyframe, respectively.

The weighting applied in the mapping loss (\cref{equation:mapping_loss}) is intended to prevent the existing map from being overwritten when it is re-observed from an incorrect pose. In this instance, we want any well-constrained portions of the map to remain largely unchanged and we want the error in the current keyframe window pose estimates to be corrected. Applying the weighting in \cref{equation:mapping_loss} with $\lambda_s>\lambda_w$ reduces the relative impact of the keyframe window views on the Gaussians, but it also has a few unintended consequences. First, the weighting reduces the gradients of the position and rotation control points that contribute to the keyframe window views. Reducing these gradients runs counter to the goal of correcting the keyframe window pose error, so we manually divide these gradients by $\lambda_w/F$, where $F$ is a set value (10 in all experiments). Second, the weighting similarly increases the gradients of the position and rotation control points that contribute to the randomly sampled keyframe views. We want these views to remain stable so we correct this by manually dividing these gradients by $\lambda_s$. Third, it has undesirable effects on the gradients of the Gaussians that are viewed primarily by the keyframe window or randomly sampled keyframes. For example, if the sets of Gaussians viewed by the keyframe window and the randomly sampled keyframes are disjoint, then the Gaussians viewed by the keyframe window will be updated too gradually, while the Gaussians seen by the randomly sampled keyframes will be updated too aggressively. We account for this by manually reweighting the gradient of each Gaussian parameter based on the fraction of the keyframes it appears in being randomly sampled versus in the keyframe window. Specifically, we divide a Gaussian's gradients by $\lambda_s (N_{\text{S}}/N_{\text{W}}) + \lambda_w (1-N_{\text{S}}/N_{\text{W}})$ where $N_{\text{S}}$ is the number of randomly sampled keyframes that view the Gaussian and $N_{\text{W}}$ is the number of keyframes in the window that view the Gaussian.

Finally, note that we blend a random background intensity with each raster $\mathbf{I}_{\mathcal{G}}(t_r)$ during mapping optimization (but not during tracking). This is done to encourage high opacities.

\section{Modified Gradient Accumulation}
\label{section:modified_gradient_accum}

In gsplat \cite{ye2025gsplat}, the image plane gradients of each Gaussian are normalized such that a single image plane gradient threshold can be consistently applied, regardless of the image resolution or the batch size of renders used per iteration. The mean of the normalized image plane gradients is computed over multiple iterations by summing them and maintaining a count of the number of rasters each Gaussian was visible in. When it comes time to apply the threshold, the mean is computed by dividing the sum by the count. Our microbolometer-aware rendering scheme complicates this because we compute multiple rasters per render and the rasters are weighted differently. To account for this, we similarly normalize the image plane gradients, but we compute the "count" differently. Instead of incrementing the count by 1 for each raster the Gaussian is visible in, we increment the count using the value of the weight matrix $\mathbf{W}_r$ associated with that raster, evaluated at the Gaussian's projected 2D image coordinate.

\section{IMU Initialization \& Update Details}
\label{section:imu_init_and_update_details}

When the IMU is initialized, the trajectory is transformed into the IMU frame with absolute scale by directly transforming the position $\mathbf{p}_i$ and rotation $\mathbf{R}_i$ control points as follows:
\begin{equation}
\begin{split}
\mathbf{p}_i &\leftarrow \mathbf{R}_i\mathbf{p}_{\text{I}}^{\text{C}} + S \mathbf{p}_i\\
\mathbf{R}_i &\leftarrow \mathbf{R}_i \mathbf{R}_{\text{I}}^{\text{C}}
\end{split}
\label{equation:imu_init_traj_transform}
\end{equation}
where $S$ is the estimated scale factor and $\mathbf{p}_{\text{I}}^{\text{C}}$ and $\mathbf{R}_{\text{I}}^{\text{C}}$ are the inverted IMU extrinsics $\mathbf{T}_{\text{C}}^{\text{I}}(t)=[\mathbf{R}_{\text{C}}^{\text{I}}, \mathbf{p}_{\text{C}}^{\text{I}}]$.

When the IMU parameters are updated, the scale update is applied by directly transforming the position control points as follows:
\begin{equation}
\begin{split}
\mathbf{p}_i &\leftarrow (\mathbf{R}_i\mathbf{p}_{\text{C}}^{\text{I}} + \mathbf{p}_i) / S_{\text{prev}} \\
\mathbf{p}_i &\leftarrow \mathbf{R}_i\mathbf{R}_{\text{C}}^{\text{I}}\mathbf{p}_{\text{I}}^{\text{C}} + S_{\text{new}} \mathbf{p}_i \\
\end{split}
\end{equation}
where $S_{\text{prev}}$ is the previous scale estimate and $S_{\text{new}}$ is the new scale estimate.

Note that applying a rotation to the rotation control points, as in \cref{equation:imu_init_traj_transform}, exactly rotates the entire spline, but the transformations made to the position control points do not exactly apply these transformations to the position spline itself. However, we have found this error to be negligible in practice. Also note that after we apply these transformations we reinitialize the trajectory and IMU bias optimizers with zeroed states. 

\section{Dataset Details}
\label{section:dataset_details}

The TRNeRF dataset \cite{specarmi_trnerf} was collected with two $640\times512$ 60 Hz thermal cameras (FLIR ADK), and a 400 Hz IMU (VectorNav VN-100). The thermal time constant of the FLIR ADK cameras is $\tau=8$ ms and the readout time per pixel is $\Delta  t_{\text{pix}}=0.043$ $\mu$s. To assess the impact of onboard noise filters, only one camera had them enabled. Unless otherwise noted, we run on the unfiltered data. We process a superset of the data tested on in TRNeRF, including 10 second, medium-speed segments at the start of each sequence that aid IMU initialization. Specifically, we run on 230, 160, and 100 seconds of data in the slow, medium, and fast sequences respectively. We obtain pseudo ground truth poses through stereo visual structure from motion, as outlined in TRNeRF, replacing COLMAP \cite{schonberger_structure--motion_2016} with GLOMAP \cite{pan2024global} and using it on 4k images per scene prior to hloc \cite{sarlin_coarse_2019}. To evaluate restoration/rendering quality, the dataset provides pseudo ground truth images for each medium and fast sequence. 

Note that the TRNeRF dataset \cite{specarmi_trnerf} is one of few thermal SLAM datasets that offer the types of aggressive camera motion we target with our method. Moreover, as highlighted in our ablation study (\cref{section:ablation_details}), it seems that onboard noise filters can significantly violate the precise camera modeling our method relies on\footnote{Note that we are specifically referring to online noise filters, not static factory calibrated or shutter based corrections. The latter category of corrections does not violate the microbolometer camera model we adopt because they are constant offsets that are absorbed into the estimated FPN. Factory calibrated corrections were enabled in the TRNeRF dataset and shutter based NUCs were performed before each recording (but not during) \cite{specarmi_trnerf}.}. These filters are enabled by default, making it likely (but unclear) that they are enabled in datasets where this setting was not mentioned. Similarly, many thermal cameras have an optional Automatic Gain Control (AGC) feature that returns 8-bit images with enhanced contrast, rather than raw 16-bit images. Like the noise filters, this contrast enhancement is a black-box algorithm that likely has spatially varying and non-linear effects that are challenging to model. Therefore, thermal datasets with AGC enabled are also unsuitable for evaluating our method. Ultimately, we test our method exclusively on the TRNeRF dataset as it is the only dataset we are aware of that accounts for these concerns.

\section{Parameter Values}
\label{section:full_parameter_values}

The full set of SLAM parameter values used for each sequence is given in \cref{table:parameter_values}. As explained in \cref{section:sparse_opt}, all spline control points are optimized with PyTorch's \texttt{SparseAdam} optimizer. For the Gaussian and bias parameters the standard \texttt{Adam} optimizer is used.

For offline refinement, the image downsample factor is set to 1, the number of rasters per render is increased to 15, the integration interval is increased to 40 ms, the accelerometer residual loss coefficient is increased to $1\times 10^{-2}$, the bias loss coefficient is set to 0, the bias learning rate is decreased to $1\times10^{-4}$, the opacity threshold is lowered to 0.005, the image plane gradient threshold is lowered to $2\times 10^{-4}$, densification and pruning is performed every 250 iterations (up to the 4,000th iteration), and a global opacity reset is performed once at 3k iterations. The gravity direction is introduced as a learned parameter and it is optimized with the standard \texttt{Adam} optimizer using a learning rate of $1\times 10^{-4}$. Offline refinement is run for a total of 15k iterations using a batch size of 10 renders.

\begin{table*}[]
\centering
\caption{SLAM parameter values.}
\label{table:parameter_values}
\resizebox{\textwidth}{!}{%
\begin{tabular}{@{}ll@{}}
\toprule
\multicolumn{1}{c}{\textbf{Description}}                                                                       & \multicolumn{1}{c}{\textbf{Value}}                                                        \\ \midrule
\multicolumn{2}{c}{\textbf{Trajectory}}                                                                                                                                                      \\ \midrule
Spline order for the positions and rotations ($k$)                                                             & 4                                                                                         \\
Knot interval for the positions and rotations ($\Delta t_c$)                                                   & $\sim 8$ milliseconds (half the frame period)                                             \\
Learning rates for the position and rotation control points                                                    & $1\times 10^{-3}, 1\times 10^{-4}$                                                        \\
Number of predicted poses used in spline fitting ($N_q$)                                                       & 10 per knot in the time range of the predictions                                                                                       \\ \midrule
\multicolumn{2}{c}{\textbf{IMU Parameters \& Loss}}                                                                                                                                           \\ \midrule
Learning rate for the gyro. and accel. biases                                                                  & $1\times 10^{-3}$                                                                         \\
Loss coefficients for the gyro. and accel. residuals ($\lambda_g$, $\lambda_a$)                                & $1\times 10^{-2}, 1\times 10^{-4}$                                                        \\
Loss coefficients for the gyro. and accel. residuals during relocalization ($\lambda_g$, $\lambda_a$)          & $5\times 10^{-4}, 5\times 10^{-6}$                                                        \\
Loss coefficient for the gyro. and accel. biases ($\lambda_{\text{bias}})$                                     & $1\times 10^{-1}$                                                                         \\ \midrule
\multicolumn{2}{c}{\textbf{Gaussians}}                                                                                                                                                        \\ \midrule
Initial Opacity                                                                                                & 0.1                                                                                       \\
Learning rates for the opacities, intensities, means, scales, and quaternions                                  & $5\times 10^{-2}, 2.5\times 10^{-3}, 1.6\times 10^{-3}, 5\times 10^{-3}, 1\times 10^{-3}$ \\ \midrule
\multicolumn{2}{c}{\textbf{Rendering}}                                                                                                                                                       \\ \midrule
Image downsample factor                                                                                        & 4                                                                                         \\
Number of rasters per render ($N_r$)                                                                           & 5                                                                                         \\
Maximum integration interval ($T_b$)                                                                           & 36 milliseconds                                                                           \\
Spline order for the pixel-wise and scalar FPN components                                                      & 2                                                                                         \\
Knot intervals for pixel-wise and scalar FPN components                                                        & 30 seconds, 5 seconds                                                                     \\
Learning rates for the pixel-wise and scalar FPN control points                                                & $1\times 10^{-4}, 1\times 10^{-4}$                                                        \\
Loss coefficient for the pixel-wise FPN component ($\lambda_{\text{FPN}})$                                     & $1\times 10^{-3}$                                                                         \\ \midrule
\multicolumn{2}{c}{\textbf{System Initialization}}                                                                                                                                           \\ \midrule
Number of Gaussians created for map initialization ($N_{\text{init}}$)                                         & 10k                                                                                       \\
Number of iterations for map initialization optimization ($N_{\text{iter}, \text{init}}$)                      & 1k                                                                                        \\ \midrule
\multicolumn{2}{c}{\textbf{Tracking \& Keyframe Selection}}                                                                                                                                  \\ \midrule
Maximum number of iterations for tracking optimization ($N_{\text{iter}, \text{track}})$                       & 100                                                                                       \\
Convergence thresholds for position and rotation                                                               & $1\times 10^{-4}$ (relative scale), $1\times 10^{-4}$ radians                             \\
Alpha threshold used to determine valid pixels in median depth computation                                     & 0.95                                                                                      \\
Translation thresholds used in selecting new keyframes (min and max)                                           & 0.02, 0.07 (both normalized by the median depth in the current frame)                     \\
Intersection over union threshold used in selecting new keyframes                                              & 0.90                                                                                      \\
Keyframe window size ($|\mathcal{W}_K|$, $N_{\text{window}}$)                                                                       & 6                                                                                         \\ \midrule
\multicolumn{2}{c}{\textbf{Mapping}}                                                                                                                                                         \\ \midrule
Maximum number of Gaussians created for a new keyframe ($N_{\text{max}}$)                                      & 5k                                                                                        \\
Alpha threshold used to determine candidate pixels for back projection in Gaussian creation                    & 0.95                                                                                      \\
Candidate pixel fraction threshold, below which Gaussian creation is skipped                                   & 0.02                                                                                      \\
Number of iterations for mapping optimization ($N_{\text{iter}, \text{map}}$)                                  & 80                                                                                        \\
Number of randomly sampled keyframes ($|\mathcal{S}_K|$)                                                       & 4                                                                                         \\
Loss coefficients for the keyframe window and randomly sampled keyframes in mapping ($\lambda_s$, $\lambda_w$) & 2, 0.33                                                                                   \\
Factor applied to the gradients of the position and rotation control points in the keyframe window             & 10                                                                                        \\
Relocalization SSIM threshold                                                                                  & 0.92                                                                                      \\ \midrule
\multicolumn{2}{c}{\textbf{Densification \& Pruning}}                                                                                                                                        \\ \midrule
Densification and pruning period                                                                               & 2k mapping iterations                                                                     \\
Condition number threshold, above which Gaussian opacities are reset                                           & 8 (applied to the natural log of the condition number)                                    \\
Opacity threshold, below which Gaussians are pruned                                                            & 0.01                                                                                      \\
Image plane gradient threshold, above which Gaussians will be split or duplicated                              & $4\times 10^{-4}$ (normalized, as described in \cref{section:modified_gradient_accum})    \\
3D scale threshold, above which Gaussians are split and below which they are duplicated                        & 0.01 (relative scale)                                                                     \\
3D scale threshold, above which Gaussians are pruned                                                           & 0.1 (relative scale)                                                                      \\ \midrule
\multicolumn{2}{c}{\textbf{IMU Initialization \& Updates}}                                                                                                                                    \\ \midrule
Time (since the first frame) at which to perform IMU initialization                                            & 10 seconds                                                                                \\
Rate at which keyframe poses are sampled for use in IMU initialization and updates                             & 2 Hz                                                                                      \\
Number of iterations for trajectory smoothing after IMU initialization                                         & 400                                                                                       \\
IMU update period                                                                                              & 10 seconds                                                                                \\ \midrule
\multicolumn{2}{c}{\textbf{Warmup}}                                                                                                                                                          \\ \midrule
Delay before B-spline extension involves fitting to predicted poses                                            & 2 keyframes                                                                               \\
Delay before motion blur and rolling shutter are modeled in rendering                                          & 20 keyframes                                                                              \\
Delay before view-diversity-based opacity resetting is performed                                               & 40 keyframes                                                                              \\
Delay before relocalization is enabled                                                                         & 150 keyframes                                                                             \\ \bottomrule
\\
\end{tabular}%
}
\end{table*}
\begin{table*}[]
\centering
\caption{Ablation study results.}
\label{table:ablation_results}
\resizebox{\textwidth}{!}{%
\begin{tabular}{lcccccccccccc}
\hline
\textbf{Ablation Method}  & \multicolumn{2}{c}{\textbf{SO}}                                                                                                         & \multicolumn{2}{c}{\textbf{MO}}                                                                                                         & \multicolumn{2}{c}{\textbf{FO}}                                                                                                         & \multicolumn{2}{c}{\textbf{SI}}                                                                                                         & \multicolumn{2}{c}{\textbf{MI}}                                                                                                         & \multicolumn{2}{c}{\textbf{FI}}                                                                                                         \\ \hline
                          & \textbf{\begin{tabular}[c]{@{}c@{}}ATE\\ (cm)\end{tabular}} & \textbf{\begin{tabular}[c]{@{}c@{}}\# of Trials\\ Completed\end{tabular}} & \textbf{\begin{tabular}[c]{@{}c@{}}ATE\\ (cm)\end{tabular}} & \textbf{\begin{tabular}[c]{@{}c@{}}\# of Trials\\ Completed\end{tabular}} & \textbf{\begin{tabular}[c]{@{}c@{}}ATE\\ (cm)\end{tabular}} & \textbf{\begin{tabular}[c]{@{}c@{}}\# of Trials\\ Completed\end{tabular}} & \textbf{\begin{tabular}[c]{@{}c@{}}ATE\\ (cm)\end{tabular}} & \textbf{\begin{tabular}[c]{@{}c@{}}\# of Trials\\ Completed\end{tabular}} & \textbf{\begin{tabular}[c]{@{}c@{}}ATE\\ (cm)\end{tabular}} & \textbf{\begin{tabular}[c]{@{}c@{}}\# of Trials\\ Completed\end{tabular}} & \textbf{\begin{tabular}[c]{@{}c@{}}ATE\\ (cm)\end{tabular}} & \textbf{\begin{tabular}[c]{@{}c@{}}\# of Trials\\ Completed\end{tabular}} \\ \cline{2-13} 
Monocular*                & 22.5                                                        & 3                                                                         & $\times$                                                    & 0                                                                         & $\times$                                                    & 0                                                                         & 7.9                                                         & 3                                                                         & $\times$                                                    & 0                                                                         & $\times$                                                    & 0                                                                         \\
No Gyro. Before IMU Init. & 51.8                                                        & 3                                                                         & 32.6                                                        & 1                                                                         & 18.0                                                        & 3                                                                         & 10.2                                                        & 3                                                                         & 8.3                                                         & 3                                                                         & $\times$                                                    & 0                                                                         \\
Gyroscope Only*           & 3.6                                                         & 3                                                                         & 5.7                                                         & 3                                                                         & 3.7                                                         & 3                                                                         & 3.3                                                         & 3                                                                         & 4.6                                                         & 1                                                                         & $\times$                                                    & 0                                                                         \\
No Roll/Blur Modeling     & 9.0                                                         & 3                                                                         & 7.3                                                         & 3                                                                         & 32.8                                                        & 1                                                                         & 4.2                                                         & 3                                                                         & 5.7                                                         & 3                                                                         & $\times$                                                    & 0                                                                         \\
No Pixel-wise FPN         & 22.7                                                        & 3                                                                         & $\times$                                                    & 0                                                                         & $\times$                                                    & 0                                                                         & 52.1                                                        & 1                                                                         & $\times$                                                    & 0                                                                         & $\times$                                                    & 0                                                                         \\
No Keyframe Weighting     & 9.0                                                         & 3                                                                         & 91.8                                                        & 3                                                                         & 3.3                                                         & 3                                                                         & 4.0                                                         & 3                                                                         & $\times$                                                    & 0                                                                         & $\times$                                                    & 0                                                                         \\
No Opacity Resetting      & 14.6                                                        & 3                                                                         & 12.8                                                        & 2                                                                         & 3.5                                                         & 3                                                                         & 6.6                                                         & 3                                                                         & 9.4                                                         & 3                                                                         & $\times$                                                    & 0                                                                         \\
No Relocalization         & 12.3                                                        & 3                                                                         & 4.8                                                         & 1                                                                         & 2.3                                                         & 3                                                                         & 6.6                                                         & 3                                                                         & 3.9                                                         & 2                                                                         & $\times$                                                    & 0                                                                         \\
With Onboard Filters      & 14.0                                                        & 3                                                                         & 4.2                                                         & 3                                                                         & 2.4                                                         & 3                                                                         & 3.7                                                         & 3                                                                         & $\times$                                                    & 0                                                                         & $\times$                                                    & 0                                                                         \\ \hline
TRGS-SLAM                 & 12.7                                                        & 3                                                                         & 5.5                                                         & 3                                                                         & 2.4                                                         & 3                                                                         & 6.4                                                         & 3                                                                         & 3.9                                                         & 3                                                                         & 4.2                                                         & 3                                                                         \\ \hline
\multicolumn{13}{l}{* Poses are relative scale and $\text{Sim}(3)$ alignment is used in evaluation.}                                                                                                                                                                                                                                                                                                                                                                                                                                                                                                                                                                                                                                                                                                                                                                                   \\
\multicolumn{13}{l}{$\times$ No trials were successfully completed.}                                                                                                                                                                                                                                                                                                                                                                                                                                                                                                                                                                                                                                                                                                                                                                                                                 
\end{tabular}%
}
\end{table*}

\section{Ablation Study}
\label{section:ablation_details}

Our ablation study results are given in \cref{table:ablation_results}. Each method tested in the ablation study was run on each sequence for three trials. We report the number of trials completed and the average RMSE ATE across the completed trials (if any). Note that some trials are incomplete because our code exits when the entire rasterized alpha image at the current keyframe is beneath the threshold used for determining pixels to use in the median depth computation (0.95 in all of our experiments). This reliably indicates severe tracking failure that has placed the current camera view in a completely unmapped area.

Running our method without an IMU (i.e., in a pure monocular mode) fails on all but the slow sequences. On the SO sequence we observed that the initial estimate of the scene structure is skewed and that this results in a globally inconsistent trajectory. We found that using the gyroscope data immediately can be pivotal to the initial estimate of structure. As shown in the results, if the gyroscope data is not used before IMU initialization, performance is significantly harmed. The gyroscope data is in fact so useful that using the gyroscope data alone (without ever initializing the IMU) achieves great results on most sequences. However, without the accelerometer data to constrain it, there is sometimes a sudden irrecoverable jump in tracking error (something that could perhaps be addressed with a dynamics prior).

Running our method without modeling motion blur or rolling shutter (i.e., using a single raster per each render) performs surprisingly well on the MO and MI sequences, though it fails on the FO and FI sequences. The learned pixel-wise FPN seems to be one of the most essential components of our method, as disabling it yields failure in most sequences and poor results in others. Keyframe weighting in mapping, view-diversity-based opacity resetting, and relocalization are important in some sequences, but not all. However, the view-diversity-based opacity resetting significantly reduces the number of Gaussians, even when it isn't pivotal to accurate tracking (e.g., in the fast outdoor sequence the final number of Guassians is $\sim$34k with view-diversity-based opacity resetting and $\sim$120k without).

The last method in the ablation study, "With Onboard Filters", corresponds to running our method on the second FLIR ADK camera in the TRNeRF dataset, which had onboard noise filters enabled. The noise filters have no apparent impact on performance in the outdoor sequences where FPN is less prominent. In the indoor sequences they have an inconsistent effect. Our method runs without issue on the SI sequence. However, on the MI and FI sequences our method fails \textit{immediately}. Note that all sequences begin with medium speed motion intended to aid IMU initialization, so the difference in performance between the SI and MI/FI sequences is not explained by their different camera speeds later in the sequences. This appears to suggest that these filters can enter a state in which they introduce significant multi-view inconsistency that violates the precise camera modeling our method relies on. As these filters are enabled by default, this may have implications to many thermal datasets collected with the same or similar cameras. 

Finally, we note that only our full proposed method is able to succeed on the FI sequence.

\section{Additional Timing Analysis}
\label{section:extended_timing}

\Cref{table:timing} gives a breakdown of timing in the MI sequence. Note that the timing data is presented hierarchically, with the indentation of the task names used to represent their parent-child relationships. Tracking runs for all 9654 frames, while mapping runs for each of the 2111 frames selected as keyframes (after system initialization). Mapping takes much longer to run than tracking, and consumes the most time overall. Within tracking, B-spline extension consumes a modest percentage of the computation time, but the bulk of the time is spent in the optimization loop. In mapping, the optimization loop completely dominates. Optimization runs for an average of 32 iterations in tracking and 83 iterations in mapping. Note that mapping is typically run for exactly 80 iterations, but is occasionally run for more when densification and pruning occurs and the iteration counter is reset. The computation time is spread quite evenly across the individual components of the optimization loops in both tracking and mapping, so there are no clear bottlenecks. 

\begin{table}[]
\centering
\caption{Timing analysis on the MI sequence.}
\label{table:timing}
\resizebox{\columnwidth}{!}{%
\begin{tabular}{@{}llllcc@{}}
\toprule
\multicolumn{4}{c}{\textbf{Task}}                                & \textbf{\begin{tabular}[c]{@{}c@{}}Avg. Time Per Run (ms)\\ $\times$ Avg. Number of\\ Runs per Parent Task\end{tabular}} & \textbf{\begin{tabular}[c]{@{}c@{}}\% of\\ Parent\\ Task\end{tabular}} \\ \midrule
\multicolumn{4}{l}{Tracking}                                     & 193.27 $\times$ 9654                                                                                                     & 36.9                                                                   \\ \midrule
   & \multicolumn{3}{l}{B-Spline Extension}                      & 27.32                                                                                                                    & 14.1                                                                   \\
   & \multicolumn{3}{l}{Optimization Iteration}                  & 5.12 $\times$ 32                                                                                                         & 84.8                                                                   \\
   &          & \multicolumn{2}{l}{Rendering}                    & 1.88                                                                                                                     & 36.8                                                                   \\
   &          &                & Pose Evaluation                 & 0.60                                                                                                                     & 31.7                                                                   \\
   &          &                & Rasterization                   & 0.57                                                                                                                     & 30.1                                                                   \\
   &          &                & Blending \& FPN                 & 0.67                                                                                                                     & 35.3                                                                   \\
   &          & \multicolumn{2}{l}{Loss Computation}             & 0.75                                                                                                                     & 14.3                                                                   \\
   &          &                & Image Loss                      & 0.08                                                                                                                     & 11.0                                                                   \\
   &          &                & IMU Loss                        & 0.64                                                                                                                     & 84.4                                                                   \\
   &          & \multicolumn{2}{l}{Backward Pass}                & 1.22                                                                                                                     & 23.1                                                                   \\
   &          & \multicolumn{2}{l}{Trajectory Optimizer Updates} & 1.05                                                                                                                     & 19.9                                                                   \\ \midrule
\multicolumn{4}{l}{Mapping}                                      & 1469.39 $\times$ 2111                                                                                                    & 61.3                                                                   \\ \midrule
   & \multicolumn{3}{l}{Gaussian Creation}                       & 0.63                                                                                                                     & 0.04                                                                   \\
   & \multicolumn{3}{l}{Optimization Iteration}                  & 17.55 $\times$ 83                                                                                                        & 99.5                                                                   \\
   &          & \multicolumn{2}{l}{Rendering}                    & 3.06                                                                                                                     & 17.4                                                                   \\
   &          &                & Pose Evaluation                 & 0.64                                                                                                                     & 20.9                                                                   \\
   &          &                & Rasterization                   & 1.63                                                                                                                     & 53.4                                                                   \\
   &          &                & Blending \& FPN                 & 0.73                                                                                                                     & 24.0                                                                   \\
   &          & \multicolumn{2}{l}{Loss Computation}             & 1.56                                                                                                                     & 8.9                                                                    \\
   &          &                & Image Loss                      & 0.20                                                                                                                     & 12.9                                                                   \\
   &          &                & FPN Loss                        & 0.35                                                                                                                     & 22.7                                                                   \\
   &          &                & IMU Loss                        & 0.98                                                                                                                     & 63.2                                                                   \\
   &          & \multicolumn{2}{l}{Backward Pass}                & 4.92                                                                                                                     & 28.1                                                                   \\
   &          & \multicolumn{2}{l}{Gradient Modification}        & 1.49                                                                                                                     & 8.5                                                                    \\
   &          & \multicolumn{2}{l}{Optimizer Updates}            & 2.90                                                                                                                     & 16.5                                                                   \\
   &          &                & Gaussian Updates                & 0.72                                                                                                                     & 24.8                                                                   \\
   &          &                & FPN Updates                     & 0.81                                                                                                                     & 27.9                                                                   \\
   &          &                & Trajectory Updates              & 1.35                                                                                                                     & 46.7                                                                   \\
   &          & \multicolumn{2}{l}{Densification}                & 2.90                                                                                                                     & 16.5                                                                   \\ \bottomrule 
\end{tabular}%
}
\end{table}

\section{Limitations}
\label{section:limits}

While our method demonstrates a unique capability to operate on the challenging TRNeRF dataset \cite{specarmi_trnerf}, it also has several limitations. Our random keyframe sampling scheme assumes a small scale scene, such that the randomly sampled keyframes will often have fields of view that overlap with the keyframe window. Similarly, our relocalization approach assumes a limited amount of drift, such that tracking is able to converge to the correct solution. These limitations could potentially be addressed by creating multiple submaps in large scale scenes, restricting the random sampling to the current submap, and performing place recognition and loop closure between submaps (similar to LoopSplat \cite{zhu2025loopsplat}). Additionally, in our testing we rescaled the 16-bit thermal images with hardcoded thresholds. In real online operation good thresholds are unlikely to be known apriori. This could be addressed by dynamically setting these thresholds and correspondingly making dynamic adjustments to the Gaussians' intensities and the learned FPN estimates. Also, in our current implementation the IMU biases are assumed constant. For long-term operation the biases would be better represented as slowly varying splines, similar to our implementation of FPN.

Additionally, our method is sensitive to features commonly enabled in thermal cameras, such as onboard noise filters and AGC. This is not an issue if these features are disabled, but it does pose a potential issue with existing datasets that had these features enabled or with thermal cameras that do not expose the option to disable them. This issue could be potentially addressed by expanding the image formation model assumed by our method to incorporate these effects. Similarly, our method does not have a strategy for dealing with shutter-based NUCs, which both interrupt the image stream and cause abrupt changes to onboard FPN corrections. However, as mentioned in \cref{section:trgs_conclusions}, it would be interesting to see if our method of estimating FPN could obviate the need for shutter-based NUCs in long-term operation. Finally, we note that these issues could also be mitigated on the manufacturer's side: if the cameras report the necessary metadata to fully reconstruct the image processing pipeline (e.g., NUC tables) these onboard operations would no longer be sources of uncertainty in camera modeling.

\end{document}